\RequirePackage{fix-cm}
\documentclass[smallextended]{svjour3}       
\smartqed  
\usepackage{graphicx}
\usepackage[square,sort,comma,numbers]{natbib}
\usepackage{tikz}
\usepackage{amsmath}
\usepackage{color}  
\usepackage[final]{changes}

\definechangesauthor[name={Ramon Morros}, color=red]{rm}
\definechangesauthor[name={Javier Ruiz}, color=blue]{jrh}
\definechangesauthor[name={Marc Maceira}, color=green]{mm}

\newcommand{\figref}[1]{Figure~\ref{#1}}
\newcommand{\equaref}[1]{Equation~\eqref{#1}}
\begin{document}

\title{3D hierarchical optimization for Multi-view depth map coding}



\author{Marc Maceira \and David Varas \and Josep-Ramon Morros \and Javier Ruiz-Hidalgo  \and Ferran Marques}


\institute{Marc Maceira Duch \at
              EDIFICI D5 DESPATX 120 C. JORDI GIRONA, 1-3 BARCELONA SPAIN \\
              Tel.: +34-93-4011627\\
              \email{marc.maceira@upc.edu} 
              \and
	David Varas \at
              EDIFICI D5 DESPATX 120 C. JORDI GIRONA, 1-3 BARCELONA SPAIN \\
              Tel.: +34-93-4011627\\
              \email{david.varas@upc.edu}                   
           \and
	Josep-Ramon Morros  \at
        EDIFICI D5 DESPATX 008 C. JORDI GIRONA, 1-3 BARCELONA SPAIN \\
        Tel.: +34-93-4015765\\
        \email{ramon.morros@upc.edu} 
           \and
	Javier Ruiz-Hidalgo  \at
        EDIFICI D5 DESPATX 008 C. JORDI GIRONA, 1-3 BARCELONA SPAIN \\
        Tel.: +34-93-4015765\\
        \email{j.ruiz@upc.edu} 
           \and
	Ferran Marques  \at
        EDIFICI D5 DESPATX 111 C. JORDI GIRONA, 1-3 BARCELONA SPAIN \\
        Tel.: +34-93-4016450\\
        \email{ferran.marques@upc.edu}
}
\date{Received: date / Accepted: date}

\maketitle
\begin{abstract}
Depth data has a widespread use since the popularity of high resolution 3D sensors. In multi-view sequences, depth information is used to supplement the color data of each view. This article proposes a joint encoding of multiple depth maps with a unique representation. Color and depth images of each view are segmented independently and combined in an optimal Rate-Distortion fashion. The resulting partitions are projected to a reference view where a coherent hierarchy for the multiple views is built. A Rate-Distortion optimization is applied to obtain the final segmentation choosing nodes of the hierarchy. The consistent segmentation is used to robustly encode depth maps of multiple views obtaining competitive results with HEVC coding standards.\\
\textbf{Available at}: http://link.springer.com/article/10.1007/s11042-017-5409-z
\end{abstract}

\section{Introduction}
\label{sec:intro}

Advances on 3D media technology and the progressively affordable 3D displays have consolidated 3D video technology in recent years.  Video applications such as 3D television and Free-viewpoint selection are able to provide depth perception and interactivity to the user. \added[id=jrh]{One way to provide this functionality is by capturing the scene from multiple viewpoints.}

Multi-view video coding systems can be classified among those that only use color data and the ones that also make use of depth data for each viewpoint. Among the ones using only color data stand out the respective extensions of the H.264~\cite{Ostermann2004} and HEVC~\cite{Sullivan2012} to multi-view environments. The multi-view video coding (MVC)~\cite{Merkle2007} was developed as the extension of H.264 standard while HEVC has the multi-view video coding extension (MV-HEVC)~\cite{Sullivan2013}. Both extensions use inter view prediction to exploit the multi-view redundancy of the N cameras employed. 

On the other hand, in the multi-view plus depth (MVD) representation, depth data of each viewpoint is encoded in addition of color data. The depth information is encoded in depth maps, where a per-pixel distance between the scene and the camera is stored. The addition of depth maps allows the rendering of virtual views in-between of the encoded camera positions (DIBR)~\cite{Fehn2004}.

The efficient compression of depth data has been studied to exploit the distinct characteristics of depth maps from standard video images. Depth maps present sharp edges separating areas with smooth transitions which are not well represented with the block-based transformations of natural video coders. The 3D extension of the high efficiency video coding (3D-HEVC)~\cite{Muller2013} presented coding tools for both color and depth maps. For depth maps, 3D-HEVC uses the similarity between color and depth data with the motion parameter inheritance. \deleted[id=rm]{Arbitrary shaped depth segments are encoded with Wedgelet and contour partitions inside the rectangular coding blocks.}

In MVD sequences, depth data associated with each view heavily augments the amount of data needed to store the 3D information. In this context, extracting a 3D model of a scene from multiple depth maps removes redundant information of the views while obtaining a unique 3D representation of the scene. \deleted[id=rm]{This representation can be further used in tasks as action detection~\cite{Zheng2015}, scene recognition~\cite{Gupta2013} or scene labeling~\cite{Wang2015}.}

\begin{figure}[ht]
    \centering{\includegraphics[trim=15mm 10mm 15mm 25mm,clip,width=\textwidth]{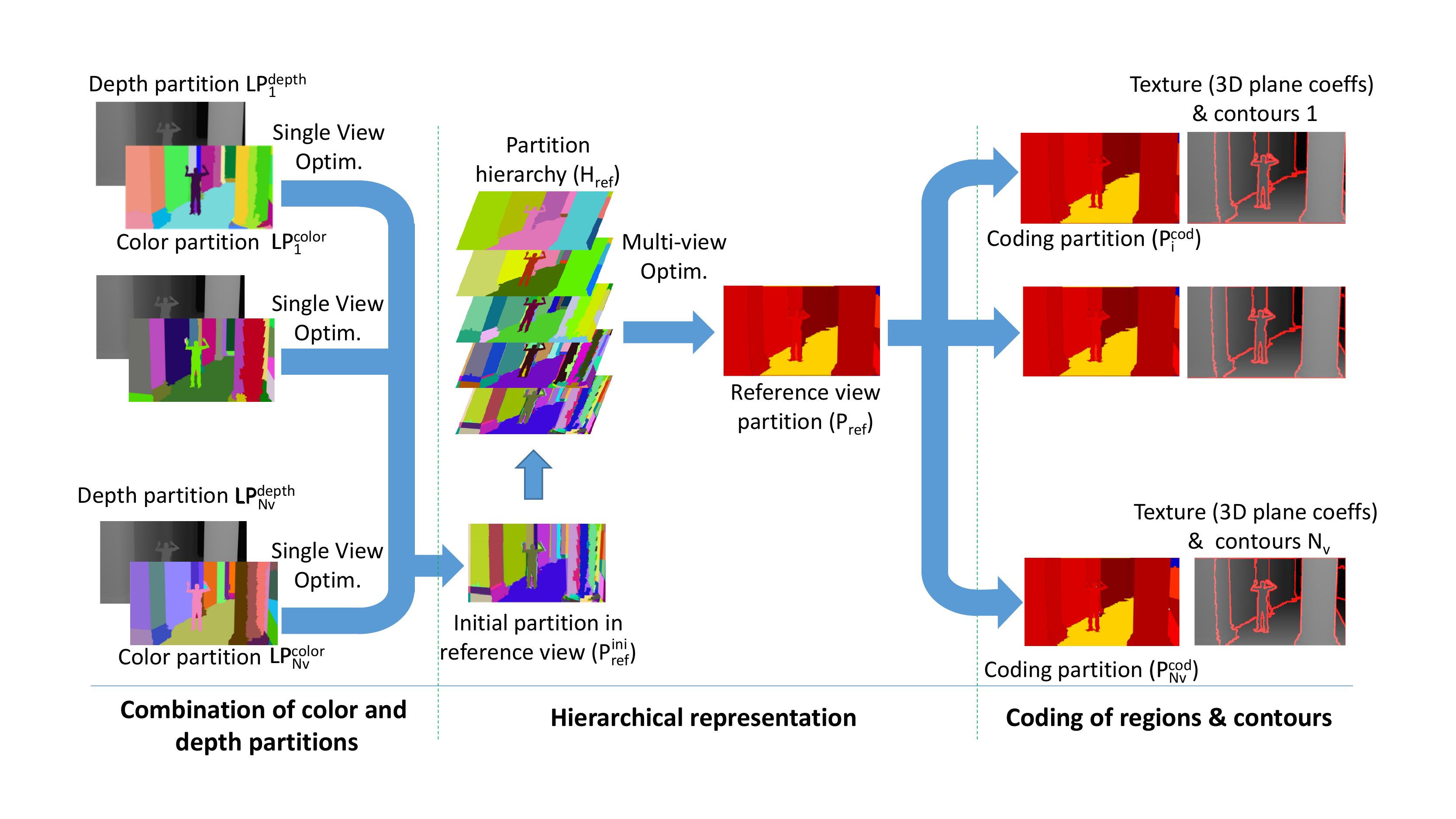}}
\caption[Multi-View optimization overview]{Overview of the multi-view optimal partition. Color \& depth map images of multiple views are  projected and combined in a reference view, where a hierarchy of regions is built (each layer shows a cut of the hierarchy). A Rate-Distortion optimization finds the optimal partition in the hierarchy which defines a coding partition in each of the input views. The coding partitions are used to encode the depth maps of each input view.}

\label{fig:fancy_scheme}
\end{figure}

\replaced[id=rm]{
In this work, we are inspired in segmentation-based coding systems~\cite{Torres1996}, where images (in our case, depth maps) are segmented into homogeneous regions; the resulting texture and contours of the partition are encoded and sent to the receiver. However, classical segmentation-based systems would use an independent segmentation of each view and the encoding process would also be performed independently for each view. We propose to build a global 3D hierarchical representation of the scene that is able to jointly model the various views in a MVD sequence. This new representation allows exploiting the spatial redundancy among views to efficiently compress the depth maps of multiple viewpoints. We start by constructing color and depth partitions for each of the views. These partitions are combined into a single hierarchical representation. The depth information of each resulting region is modeled as a 3D plane, thus providing a global 3D hierarchical scene representation. This 3D planar representation is a convenient model for the smooth regions with sharp transitions in depth maps~\cite{Muller2011}. As long as the sharp edges in depth maps lay in-between regions, the planar models can approximate the texture information compactly.

A Rate-Distortion optimization process over the previous hierarchy (hierarchical optimization) allows obtaining an optimal (for a given bit budget) coding partition for each one of the different views automatically. Depth map coding is then performed by sending the plane coefficients for all regions of the view's coding partitions, as well as the contours of these partitions. 
This set of partitions is consistent (regions representing the same objects in the different views are given the same labels) because the cameras parameters are used through all the process. 
Having a global multiview model implies that corresponding regions from different views can be represented with a unique planar model, thus reducing the number of bits necessary. \figref{fig:fancy_scheme} shows a graphical overview of the proposed multi-view representation and coding scheme.
}{In this work, we propose to use the spatial redundancy among views to compress the depth maps of multiple viewpoints. The proposed scheme obtains a consistent segmentation of the scene from multiple depth maps (set of coherent 2D partitions) and a unique 3D planar representation (2D regions are modeled with a 3D plane).}

\deleted[id=jrh]{In our proposed scheme, a region-based representation models the 3D scene. Starting from an image partition, each region is represented with a 3D planar model. 
Since the depth information can be projected from one viewpoint to another using the camera parameters, regions from different views can be related. Once regions from multiple views are related, a single 3D plane is fitted using the 3D projected points from those regions. Regions from different views are represented with a unique model. The number of regions in the 3D representation depends on the available bit budget and is determined by an optimization process over a hierarchy of regions.
}

\deleted[id=rm]{Furthermore, we combine color and depth partitions to efficiently encode the depth information.} One of the problems of segmentation based coding systems is the high cost of sending the contours. \replaced[id=rm]{In this work, we will assume that color images are already encoded and available at the decoder. Thus}{Assuming that the color images are already encoded}, depth boundaries can be approximated by a color partition without cost, because the color partition can be reproduced from the color image at the decoder. \replaced[id=rm]{Only the approximation error (depth contours not present in the color image) will need to be explicitly sent, at a reduced coding cost.}{Depth boundaries are able to increase the quality of the depth map with the added cost of explicitly encoding the location of the depth edges.}

\deleted[id=mm]{
Using the hierarchical representation we are able to retrieve the optimal combination between color and depth edges.}


The main contributions of this paper are:
\begin{itemize}
  \item A consistent segmentation of the multiple views and a hierarchical representation of the different views.
  \item A new global scene representation based on a 3D planar model obtained from a hierarchical optimization. 
  \item A method to efficiently encode the depth maps from multiple views, based on the global scene representation.
\end{itemize}

The outline of the paper is as follows. Related work in 3D planar representations and multi-view depth coding is studied in Section~\ref{sec:related}. In Section~\ref{sec:rdHierOpt}, the generic hierarchical optimization method is presented. In Section~\ref{sec:mv3Drepres}, the consistent multi-view depth map representation is introduced. Details of encoding the information of the multiple views are found in Section~\ref{sec:multi-view-coding}. Finally, we discuss the experimentation results and conclusion in Section~\ref{sec:results} and Section~\ref{sec:conclusion} respectively.

\section{Related Work}
\label{sec:related}

The use of 3D planar models to represent the structure of a scene has been explored in multiple applications. 3D planar models have shown to be useful to extract the structure of a scene using only color data in a stereo configuration~\cite{Sinha2009}, to co-segment multiple view objects~\cite{Kowdle2012}, to recover the structure of the scene from panoramic sequences~\cite{Micusik2009} or from a multi-camera environment~\cite{Yin2015}. 
Multiple planes are detected and tracked in Time of Flight depth images using a layered scene decomposition in~\cite{Schwarz2011}.

More similar to our configuration, in~\cite{Barrera2014} the 3D point cloud from multiple RGB-D cameras is projected to the 3D world and used to generate a set of 3D planar patches consistent among the views. This patches are obtained with a Markov random fields using a voxelization of the scene and several 3D planar candidates. More recently, in~\cite{Verleysen2016} the same problem is tackled in a wide-baseline stereo configuration. Each image is segmented independently and a fitting process assigns a 3D plane to each region. Those planes are used as candidates in an energy-minimization problem, which optimizes the error of the model and the smoothness over neighboring regions. 

Segmenting point clouds coming from RGB-D sensors has been tackled as a part of semantic labeling of indoor scenes and scene understanding. In~\cite{Silberman2012} a hierarchical segmentation is proposed using depth clues to infer the relations between objects and using priors to classify the different objects. In~\cite{Ren2012} an initial superpixel segmentation is used to compute kernel descriptors such as gradient, color and surface normal to build a hierarchy that is used to assign the labels to each node using markov random field. Gupta et al.~\cite{Gupta2013} generalize the gPb-ucm hierarchical segmentation to incorporate depth information for semantic segmentation. Lately, in~\cite{Wang2015} an unsupervised framework where joint feature learning and encoding is proposed for RGB-D scene labeling.


Planar models have been also proposed for depth representation on 3DTV applications. In~\cite{Ozkalayci2014} a Markov random field model that mimics a Rate-Distortion trade-off is used in a stereo configuration to obtain a co-segmentation with planar approximations. A single reference MVD format is also proposed to fuse the information of the stereo views in a single reference. 

Some depth map coding techniques propose improvements of the depth coding modes of 3D-HEVC. In~\cite{Lucas2015} an intra prediction framework with a flexible block partition scheme. Depth edges that cannot be predicted are explicitly encoded. Similarly, in~\cite{Merkle2016} a complete overhaul of the 3D-HEVC is defined. It incorporates planar signals to represent the smooth changes of flat scene areas, a new intra-picture prediction for this model and an improved wedgelet segmentation.


A joint color and depth coding for multi-view video is proposed in~\cite{Gao2016}. Making use of DIBR, the color and depth information is projected obtaining a inter-view prediction. This inter-view prediction has missing pixels which are inpainted using minimum explicitly encoding.

In~\cite{Maceira2016} we proposed a region based coding of depth maps for a single-view. There we defined depth hierarchies to segment a depth map from one RGB-D image but only a flat partition was explored for coding. \deleted[id=rm]{A planar model for each region was used to build a hierarchical depth segmentation which complements a color segmentation.} 
\replaced[id=rm]{In the current paper we extend the depth hierarchies defined in~\cite{Maceira2016} to deal with multiple views on an optimal fashion.}{In the current paper we use the depth hierarchies defined in~\cite{Maceira2016} exploiting the richness of the hierarchical representation. The final coding regions are obtained combining both partitions in an optimal fashion. This 3D planar representation is a convenient model for the smooth regions with sharp transitions in depth maps~\cite{Muller2011}. As long as the sharp edges in depth maps lay in-between regions, the planar models can approximate the texture information compactly.}

\section{Rate-Distortion hierarchy optimization }
\label{sec:rdHierOpt}
In this Section, we define a generic methodology to find the optimal partition in a hierarchy in terms of Rate-Distortion. The notation used is consistent with~\cite{Varas2015}, where an optimization on multiple hierarchies is used for semantic segmentation of sequences. 
\added[id=mm]{
The optimization presents two benefits. Firstly, the encoding parameters are selected optimally for each region and secondly, 
the optimal final number of regions needed to encode the depth map can be automatically obtained. In this work, we will use this Rate-Distortion optimization in various steps of the proposed algorithm.}


\deleted[id=rm]{Inspired by~\cite{Charikar2003,Glasner2011,Varas2015}, we propose to solve the Rate-Distortion optimization problem as a quadratic semi-assignment problem (QSAP), restricting the solution to nodes of the hierarchy. Partitions are defined in terms of boundaries between regions and hierarchical constraints are added to solve the QSAP problem with a linear programming relaxation approach.
The QSAP approach have two main advantages over other common approaches such as the dynamic programming solution. Firstly, QSAP defines the relation between regions (whether they are merged or not) in terms of their common contour. This relation allows to represent the contour cost of the union of two regions easily. Secondly, QSAP can be generalized to work with multiple hierarchies, which would allow applying the procedure defined in this work to relate frames of multiple time instants in order to remove temporal redundancy.}

\subsection{Hierarchy Definition}
\label{sec:hierarchy_def}

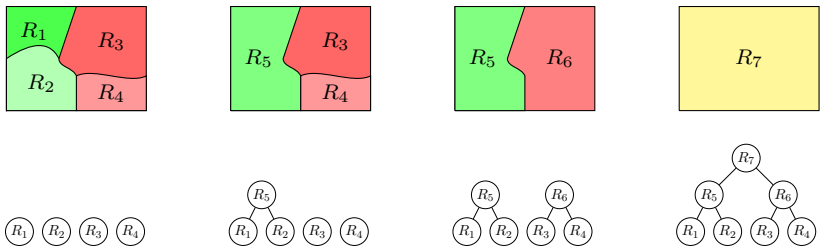
\begin{figure}[ht]
\mbox{\footnotesize\begin{minipage}[t]{.24\linewidth}%
\centering
\begin{tikzpicture}[scale=0.46]
\filldraw[fill=red!60!white, draw=black] (0,0) -- (4,0) -- (4,3) -- (0,3) -- (0,0);
\filldraw[fill=green!70!white, draw=black] (2,3) -- (1.5,1.5)--(0,0)--(0,3)--(2,3);
\filldraw[fill=red!40!white, draw=black] (2,1) .. controls (2.5,1.25) and (3.5,0.75) .. (4,1) -- (4,0) -- (2,0) -- (2,1);
\filldraw[fill=green!30!white, draw=black] (0,1.5) .. controls (0.75,2) and (1.25,2) .. (1.5,1.5) .. controls (1.5,1.25) and (1.9,1.25) .. (2,1) -- (2,0) -- (0,0)--(0,1.5);
\draw (0.8,2.3) node {$R_1$};
\draw (3,2) node {$R_3$};
\draw (1,0.8) node {$R_2$};
\draw (3,0.5) node {$R_4$};
\end{tikzpicture}
\end{minipage}
\begin{minipage}[t]{.24\linewidth}
\centering
\begin{tikzpicture}[scale=0.46]
\filldraw[fill=red!60!white, draw=black] (0,0) -- (4,0) -- (4,3) -- (0,3) -- (0,0);
\filldraw[fill=red!40!white, draw=black] (2,1) .. controls (2.5,1.25) and (3.5,0.75) .. (4,1) -- (4,0) -- (2,0) -- (2,1);
\filldraw[fill=green!50!white, draw=black] (0,0)--(0,3)-- (2,3) -- (1.5,1.5) .. controls (1.5,1.25) and (1.9,1.25) .. (2,1) -- (2,0) -- (0,0);
\draw (0.8,1.5) node {$R_5$};
\draw (3,2) node {$R_3$};
\draw (3,0.5) node {$R_4$};
\end{tikzpicture}
\end{minipage}
\begin{minipage}[t]{.24\linewidth}
\centering
\begin{tikzpicture}[scale=0.46]
\filldraw[fill=red!50!white, draw=black] (0,0) -- (4,0) -- (4,3) -- (0,3) -- (0,0);
\filldraw[fill=green!50!white, draw=black] (0,0)--(0,3)-- (2,3) -- (1.5,1.5) .. controls (1.5,1.25) and (1.9,1.25) .. (2,1) -- (2,0) -- (0,0);
\draw (0.8,1.5) node {$R_5$};
\draw (3,1.5) node {$R_6$};
\end{tikzpicture}
\end{minipage}
\begin{minipage}[t]{.24\linewidth}
\centering
\begin{tikzpicture}[scale=0.46]
\filldraw[fill=yellow!50!white, draw=black] (0,0) -- (4,0) -- (4,3) -- (0,3) -- (0,0);
\draw (2,1.5) node {$R_7$};
\end{tikzpicture}
\end{minipage}}\\[4mm]
\mbox{\large\begin{minipage}[b]{.237\linewidth}
\centering
\scalebox{0.5}{\begin{tikzpicture}[scale=0.65,level/.style={sibling distance=30mm/#1},every node/.style={inner sep=1.8pt,circle,draw}]
\node [draw=none] {}
  child {node [draw=none] {} edge from parent[draw=none]
    child {node {$R_1$} edge from parent[draw=none]}
    child {node {$R_2$} edge from parent[draw=none]}
    }
  child {node [draw=none] {} edge from parent[draw=none]
    child {node {$R_3$} edge from parent[draw=none]}
    child {node {$R_4$} edge from parent[draw=none]}
    };
\end{tikzpicture}}
\end{minipage}
\begin{minipage}[b]{.237\linewidth}
\centering
\scalebox{0.5}{\begin{tikzpicture}[scale=0.65,level/.style={sibling distance=30mm/#1},every node/.style={inner sep=1.8pt,circle,draw}]
\node [draw=none]{}
  child {node  {$R_5$} edge from parent[draw=none]
    child {node (b) {$R_1$}}
    child {node (g) {$R_2$}}
    }
  child {node [draw=none] {} edge from parent[draw=none]
    child {node [circle,draw] {$R_3$} edge from parent[draw=none]}
    child {node [circle,draw] {$R_4$} edge from parent[draw=none]}
    };
\end{tikzpicture}}
\end{minipage}
\begin{minipage}[b]{.237\linewidth}
\centering
\scalebox{0.5}{\begin{tikzpicture}[scale=0.65,level/.style={sibling distance=30mm/#1},every node/.style={inner sep=1.8pt,circle,draw}]
\node [draw=none]{}
  child {node {$R_5$} edge from parent[draw=none]
    child {node (b) {$R_1$}}
    child {node (g) {$R_2$}}
    }
  child {node  {$R_6$} edge from parent[draw=none]
    child {node (b) {$R_3$}}
    child {node (g) {$R_4$}}
    };
\end{tikzpicture}}
\end{minipage}
\begin{minipage}[b]{.237\linewidth}
\centering
\scalebox{0.5}{\begin{tikzpicture}[scale=0.65,level/.style={sibling distance=30mm/#1},every node/.style={inner sep=1.8pt,circle,draw}]
\node (z){$R_7$}
  child {node (a) {$R_5$}
    child {node (b) {$R_1$}}
    child {node (g) {$R_2$}}
    }
  child {node (a) {$R_6$}
    child {node (b) {$R_3$}}
    child {node (g) {$R_4$}}
    };
\end{tikzpicture}}
\end{minipage}
}

\caption{Hierarchical representation of an image. At each step the two most similar regions are merged.}
\label{fig:build_bpt}
\end{figure}

The representation of a hierarchy can be depicted with nodes as shown in \figref{fig:build_bpt}. Each node of the hierarchy represents a region in the image, and the parent node of a set of regions represents their merging. In all stages we assume that this hierarchy is binary (regions are merged by pairs). This structure is named as Binary Partition Tree~\cite{Salembier2000}. Commonly, such hierarchies are created using a greedy region merging algorithm that, starting from an initial leaf partition $LP$, iteratively merges the most similar pair of neighboring regions according to a region similarity criteria.

The merging process ends when the whole image is represented by a single region, which is the root of the tree. The set of mergings that creates the tree, from the leaves to the root, is denoted as merging sequence (\figref{fig:build_bpt}). In a binary hierarchy, a merging sequence contains N partitions, where $N$ is the number of leaves (regions in $LP$). This is the set of partitions that is usually analyzed when working with hierarchies. Still, once the hierarchy is built, an analysis on the whole hierarchy could obtain partitions which are not included in the merging sequence  \added[id=rm]{(for instance, merging regions 3 and 5 in the example of \figref{fig:build_bpt})}

\subsection{Rate-Distortion Problem Definition}
\label{sec:hierarchy_rd}

\deleted[id=rm]{Nowadays, it is widely accepted that multiresolution region-based descriptions provide a rich framework for image and video analysis~\cite{Arbelaez2014}. 
In general, this leads to define a partition from the hierarchy, however a partition should be selected within the hierarchy. We propose to select it with a Rate-Distortion optimization.}

The Rate-Distortion problem consists in finding the optimal coding that minimizes the distortion $D$ of the image with the constraint that the total cost $R$ is below a given budget~\cite{Ortega98}. 
In our case, this optimal coding consists in finding the optimal partition to describe the image. 

In a hierarchical representation, the Rate-Distortion function has to be defined (see~\cite{Ortega98}) with $R$ and $D$ additive measures in the hierarchy $H$:

\begin{equation}
R_{H} = \sum_{leaves} R_k
\end{equation}
\begin{equation}
D_{H} = \sum_{leaves} D_k
\end{equation}

Each node of the hierarchy can be encoded using a set of coding techniques generating a set of $R_k$ and $D_k$ for each region $k$. In this work, the distortion measure $D_{k}$ is computed as the Square Error between the estimated texture values and the real ones:

\begin{equation} 
D_{k} = \sum\limits_{n=1}^{N_{k}}|\hat{g}(n) - g(n)|^2
\label{eq:D}
\end{equation}

where $N_{k}$ is the number of pixels of the region $k$, $g$ is the texture value of the pixels of the region and $\hat{g}$ is the estimated values with the texture coding option.

Typically, in region-based coding, two terms are needed to encode the region. The cost to store the position of the contours of the region $(R_k^{C})$ and the cost to store the texture coefficients $(R_k^{T})$. Hence, for each region their rate cost $R_{k}$ is defined as:

\begin{equation} 
R_{k} = R_k^{C} + R_k^{T}
\label{eq:R}
\end{equation}

The constrained problem can be converted into an equivalent unconstrained one by using Lagrangian relaxation~\cite{Ortega98}. Cost and distortion are combined using a positive 
multiplier $\lambda$ that defines the trade-off between the distortion allowed and the number of bits spent:

\begin{equation}
J_k = D_k +\lambda R_k
\end{equation}

The best partition can be obtained by using a bottom-up local analysis at each node:

\begin{equation} 
J_{parent} \leq J_{ch_1} + J_{ch_2}
\label{eq:local_analysis}
\end{equation}
where $J_{ch}$ is the cost of the cheapest path under the child node.

\subsection{Hierarchy Optimization}
\label{sec:hierarchy_opt}

The objective of the hierarchy optimization is to find the optimal boundary configuration that defines a partition using nodes from the hierarchy $H$ using the previously defined Rate-Distortion constraints. 

\added[id=rm]{Inspired by~\cite{Charikar2003,Glasner2011,Varas2015}, we propose to solve the Rate-Distortion optimization problem as a quadratic semi-assignment problem (QSAP), restricting the solution to nodes of the hierarchy. Partitions are defined in terms of boundaries between regions and hierarchical constraints are added to solve the QSAP problem with a linear programming relaxation approach. The QSAP approach have two main advantages over other common approaches such as the dynamic programming solution. Firstly, QSAP defines the relation between regions (whether they are merged or not) in terms of their common contour. This relation allows to represent the contour cost of the union of two regions easily. Secondly, QSAP can be generalized to work with multiple hierarchies, which would allow applying the procedure defined in this work to relate frames of multiple time instants in order to remove temporal redundancy.}

This optimization can be stated as a constrained minimization problem:

\begin{align}
& \min_{B}\ tr(QB) = \min_{B}\ \sum_{m,n}q_{m,n}b_{m,n} \label{eq:min1} \\ 
& s.t.\quad b_{m,n}\in\{0,1\}\ \forall m, n\ b_{m,m}=0 \nonumber
\end{align}
where matrix $Q$ is an affinity matrix that measures the quality of all the possible partitions in the hierarchy $H$. Matrix $B$ encodes those partitions as a binary matrix, where elements $b_{m,n}=1$ if the boundary between leaves $m$ and $n$ is active (regions $m$ and $n$ have not been merged) and $b_{m,n}=0$ otherwise. 

In our case, the elements in matrix $Q$ can be defined to mimic the Rate-Distortion Lagrangian decision as:

\begin{equation}
q_{ch_1,ch_2} = J_{parent} - (J_{ch_1} + J_{ch_2})
\label{eq:q}
\end{equation}

\begin{figure}[h]
\mbox{\footnotesize\begin{minipage}[t]{.5\linewidth}%
\centering
\scalebox{0.9}{\begin{tikzpicture}[scale=0.65,level/.style={sibling distance=30mm/#1},every node/.style={inner sep=1.8pt,circle,draw}]
\node (z){$R_7$}
  child {node[fill=green!50!white] (a) {$R_5$}
    child {node (b) {$R_1$}}
    child {node (g) {$R_2$}}
    }
  child {node (a) {$R_6$}
    child {node[fill=red!60!white] (b) {$R_3$}}
    child {node[fill=red!40!white] (g) {$R_4$}}
    };
\end{tikzpicture}}
\end{minipage}
\begin{minipage}[b]{.5\linewidth}
\centering
\begin{tikzpicture}[scale=0.7]
\filldraw[fill=red!60!white, draw=black] (0,0) -- (4,0) -- (4,3) -- (0,3) -- (0,0);
\filldraw[fill=red!40!white, draw=black] (2,1) .. controls (2.5,1.25) and (3.5,0.75) .. (4,1) -- (4,0) -- (2,0) -- (2,1);
\filldraw[fill=green!50!white, draw=black] (0,0)--(0,3)-- (2,3) -- (1.5,1.5) .. controls (1.5,1.25) and (1.9,1.25) .. (2,1) -- (2,0) -- (0,0);
\draw (0.8,1.5) node {$R_5$};
\draw (3,2) node {$R_3$};
\draw (3,0.5) node {$R_4$};
\end{tikzpicture}
\end{minipage}
}
\caption{Example of Lagrangian optimization over the hierarchy presented in \figref{fig:build_bpt}. The optimization process activates the nodes with minimum Lagrangian $J$ (regions 3, 4 and 5), thus finding the best partition for a given $\lambda$. \deleted[id=rm]{and the best coding technique for each region.} }
\label{fig:lagrang_bpt}
\end{figure}
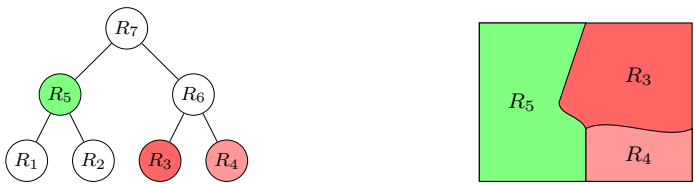

This allows to introduce the local node analysis of \equaref{eq:local_analysis} into the global framework optimization of \equaref{eq:min1}. The idea is that
nodes with minimum Lagrangian $J$ are favored, thus leading to the optimal partition for a given $\lambda$. \replaced[id=rm]{\figref{fig:lagrang_bpt} presents an example of Lagrangian optimization over the hierarchy presented in \figref{fig:build_bpt}. The algorithm examines all the nodes in the hierarchy (nodes 1-7) and selects the ones that present a minimum Lagrangian (nodes 3, 4 and 5). The final partition (shown in the right side of the figure) is composed of the regions represented by these nodes.}{(see \figref{fig:lagrang_bpt}).}

The hierarchy $H$ contributes in two aspects to the optimization process. Firstly, it defines the mergings between regions of its leaves partition $LP$ to form clusters. Secondly, it also includes the order in which these regions should be merged to represent each node of the tree. As in~\cite{Varas2015}, these restrictions are encoded using two coupled constraints per node. The first constraint imposes that all the variables representing boundaries between two siblings should have the same value. The second constraint imposes that for a given node, a variable representing a boundary between two siblings can only impose a merging if all the leaves associated with the node are merged.

The first constraint is defined as: 

\begin{equation} \sum_{n\neq l}^{m,n}b_{m,n}=(N_{c}-1)b_{m,l} \label{eq:c1} \end{equation}

where $N_c$ is the total number of common region boundaries from the leaf partition that represents the union of both siblings, $m$ is a region from the first sibling and $n,l$ are regions from the second sibling. 

The second as:

\begin{equation} \sum^{n,l}b_{n,l}\leq N_{m}b_{m,o} \label{eq:c2} \end{equation}

where $N_m$ is the total number of inner region boundaries from the leaves partition of both siblings, $m$ and $n$ are regions from the first sibling and $n,l$ are regions from the second sibling. 


Adding these two constraints to the optimization process of \equaref{eq:min1} results in:

\begin{gather}
\min_{B}\sum_{m,n}q_{m,n}b_{m,n} \label{eq:min2} \\ 
s.t.\quad b_{m,n}\in \{0,1\}\quad b_{m,n}=b_{n,m} \quad \forall n,m \nonumber \\ 
\sum_{n\neq l}^{m,n} b_{m,n}=(N_{c}-1)b_{m,l}, \quad \sum^{n,l}b_{n,l}\leq N_{m} b_{m,o} \quad \forall p\in \{H\} \nonumber
\end{gather}

where $p$ represents any parent node in the collection of hierarchies. The result of this optimization is a binary matrix $B^*$ that describes the optimal partition $\{P^*(\lambda)\}$. Varying $\lambda$, different optimal Rate-Distortion partitions are found. 
\added[id=rm]{Each $\lambda$ corresponds to a different potential solution on the convex hull of the operational Rate-Distortion points. It can be demonstrated that the set of solutions is finite~\cite{Shoham1988}. Because of the monotonicity of $R(\lambda)$, only a small subset of $\lambda$ need to be searched and the convergence of the algorithm is ensured in all cases.}
\deleted[id=rm]{Once the final partition is found, the vector $b$ that encodes the active boundaries between leaves is stored. This information is needed to recover the final coding partition from the leaves partition.}

\section{Multi-view 3D Representation}\label{sec:mv3Drepres}

\begin{figure*}[h]
    \centering
  \includegraphics[trim=0 120 0 160,clip,width=\textwidth]{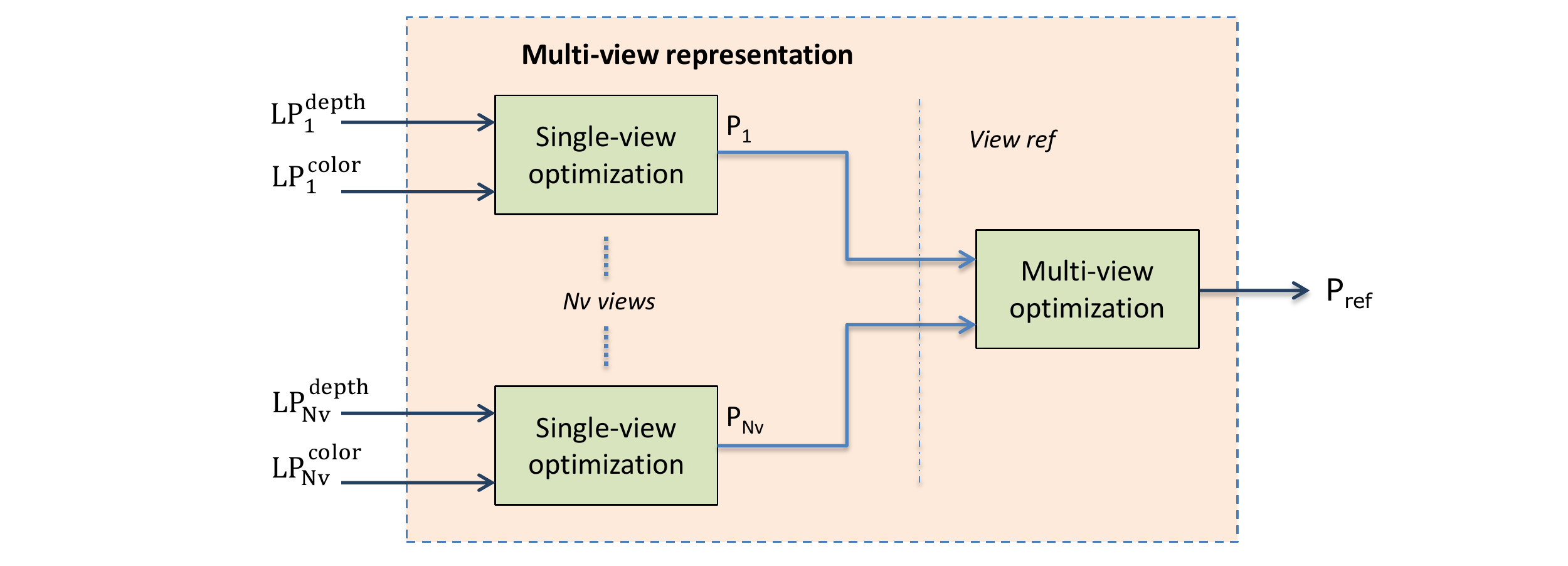}
    \caption{Multi-view representation. For each view $i$, a joint color and depth optimization finds a $P_i$ partition. The $N_v$ $P_i$ partitions are projected to the reference view where a multi-view optimization finds the best set of regions for each partition.}
    \label{fig:encoder-scheme}
\end{figure*}

The proposed multi-view depth map coding system fuses the information of multiple views in a unique 3D representation. Our method builds a 3D representation from a set of initial 2D partitions for each of the views. 
The global scheme is depicted in \figref{fig:encoder-scheme}. It starts by finding a partition for each individual view $i$ which merges color and depth partitions optimally, reducing the computational complexity of the multi-view process. The single-view processing is done independently for each view as shown in \figref{fig:1view-opt-scheme} and explained in Section~\ref{sec:base-view-opt}. 

The information of the multiple views is accumulated in the reference view $view_{ref}$ by projecting the views partitions into $view_{ref}$. Then, an optimization process (depicted in \figref{fig:multi-view-opt-scheme}) obtains $P_{ref}$ which defines the final coding partitions for each view. Complete details on this process are given at Section~\ref{sec:multi-view-opt}. 

We refer the optimization process of \equaref{eq:min2} as:

\begin{equation}
P^{*}(\lambda) = Opt_{\lambda}(H)
\end{equation}

where for a given $\lambda$ an optimal partition $P^{*}$ is extracted from $H$. For simplicity, we will reduce the notation of optimal partitions from $P^*(\lambda)$ to $P$.

In this work we use the initial color and depth leaves partitions from~\cite{Maceira2016}. We name those initial partitions as $LP^{color}_{i}$ and $LP^{depth}_{i}$ respectively.
Hierarchies in the single-view and the multi-view stages are built as in~\cite{Maceira2016}: region contents are projected into 3D and modeled as 3D planes with RANSAC~\cite{Fischler1981}. The merging criterion is based on 3D plane similarity.  Both stages find optimal partitions $P^{*}$ inside the hierarchy with the method presented in Section~\ref{sec:rdHierOpt}.



\subsection{Single-View Optimization} 
\label{sec:base-view-opt}

\begin{figure}[ht]
    \centering    \includegraphics[trim=0 180 0 120,clip,width=\columnwidth]{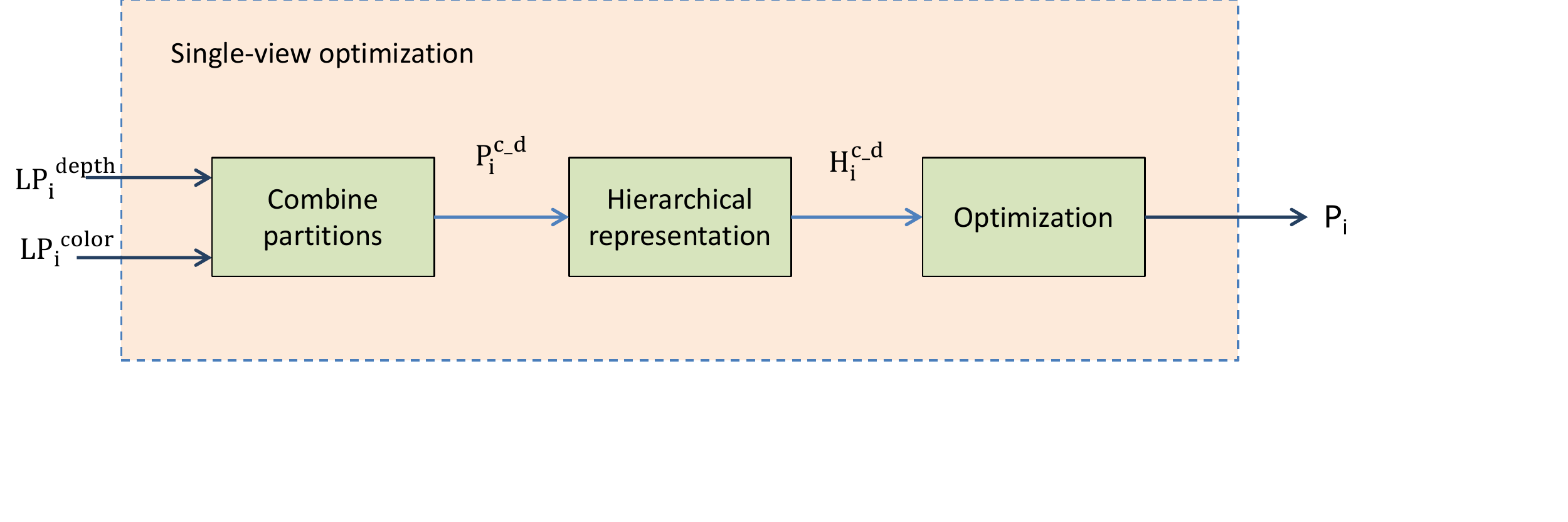}
    \caption[Multi-View representation: single-view optimization]{Optimization process for each view. A combined color and depth optimization is performed finding an optimal combined partition.}
    \label{fig:1view-opt-scheme}
\end{figure}

In the Single-View Optimization $LP^{color}_{i}$ and $LP^{depth}_{i}$ are combined into an optimal partition $P_i$ for each view $i$ as shown in \figref{fig:1view-opt-scheme}. 


$LP^{color}_{i}$ and $LP^{depth}_{i}$ are fused into a partition $P^{c\_d}_{i} = LP^{color}_{i} \bigcap LP^{depth}_{i}$ that preserves all the boundaries from both partitions. The $P^{c\_d}_{i}$ is used to build a hierarchy $H^{c\_d}_{i}$. The optimization procedure introduced in Section~\ref{sec:hierarchy_rd} over that hierarchy obtains an optimal Rate-Distortion partition $P_{i}$:

\begin{equation} 
P_{i} = Opt_{\lambda}(H^{c\_d})
\end{equation}

For each view $i$, \equaref{eq:q} can be expressed as follows:
\begin{gather}
q_{ch_{i,1},ch_{i,2}} = J_{p_i} - (J_{ch_{i,1}} + J_{ch_{i,2}}) = \label{eq:q_follow}\\
(D_{p_i} + \lambda R_{p_i}) - (D_{ch_{i,1}} + \lambda R_{ch_{i,1}}) - (D_{ch_{i,2}} + \lambda R_{ch_{i,2}}) = \nonumber \\
\triangle D_{ch_{i,1},ch_{i,2}} + \lambda \triangle R_{ch_{i,1},ch_{i,2}} \nonumber
\end{gather}
where $p_i$ is the parent region resulting of the union of children $ch_{i,1}$ and $ch_{i,2}$:  
\begin{equation} 
p_i = ch_{i,1} \bigcup ch_{i,2} 
\label{eq:union}
\end{equation}
$\triangle D_{ch_{i,1},ch_{i,2}}$ is the increment in distortion caused by the union of \equaref{eq:union} while $\triangle R_{ch_{i,1},ch_{i,2}}$ is the increment in rate of this union. Typically, the $\triangle D_{ch_{i,1},ch_{i,2}}$ is positive as representing the depth map with less regions will increase the distortion and $\triangle R_{ch_{i,1},ch_{i,2}}$ is negative.

\begin{figure}[ht]
    \centering    \includegraphics[trim=0 160 0 120,clip,width=0.9\columnwidth]{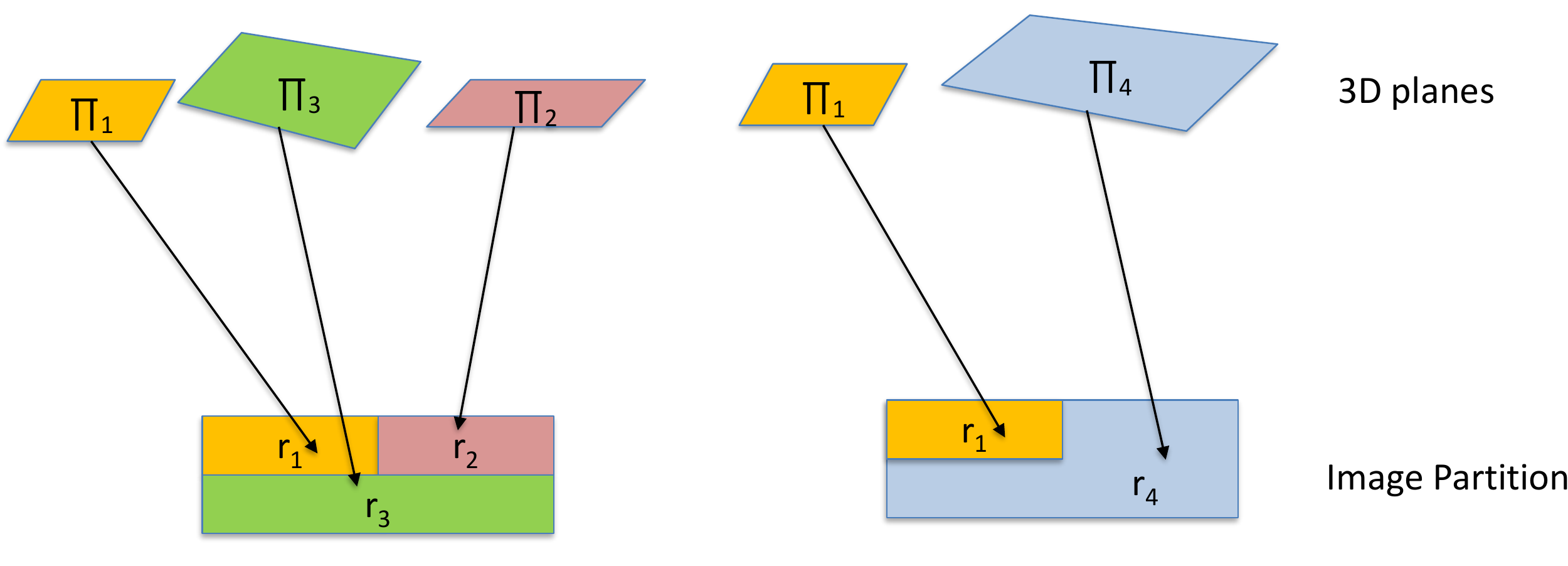}
    \caption[Multi-View representation: single-view distortion]{The Single-View distortion is computed projecting each 3D planar model to all the pixels of the corresponding region. The $r_4$ distortion generated with $\Pi_4$ is compared with a representation using the child regions $r_2$ and $r_3$, represented with models $\Pi_2$ and $\Pi_3$.}
    \label{fig:1v_dist_planes}
\end{figure}

The distortion of each region $k$ of the hierarchy is computed as the square error between the 3D planar model and the pixel values of the region, particularizing \equaref{eq:D} as:

\begin{equation} 
D_{i,k} = \sum\limits_{n=1}^{N_{i,k}^{Pix}} | Proj_i(\Pi_k,n) - g_i(n) |^2
\label{eq:Di}
\end{equation}

where $\Pi^{3D}_k$ is the plane model for region $k$, $N_{i,k}^{Pix}$ is the number of pixels of region $k$, $Proj_i(\Pi_k,n)$ is the value of the projected 3D planar model to the pixel $n$ of region $k$ and $g_i(n)$ is the depth value of pixel $n$ for each view $i$. An example of the projection step is shown in \figref{fig:1v_dist_planes}.

The $\triangle D_{ch_{i,1},ch_{i,2}}$ is computed as:
\begin{equation} 
\triangle D_{ch_{i,1},ch_{i,2}} = D_{p_i} - D_{ch_{i,1}} - D_{ch_{i,2}} \label{eq:increment_Di}
\end{equation}

To compute the $\triangle R_{ch_{i,1},ch_{i,2}}$ two terms are present: texture and contour. For texture:

\begin{equation} 
\triangle R^{T}_{ch_{i,1},ch_{i,2}} = R^T_{p_i} - R^T_{ch_{i,1}} - R^T_{ch_{i,2}} = -R^T
\label{eq:increment_Ri}
\end{equation}
since the cost of texture is constant ($R^T$) for all regions.


Contours of a region may come from two partitions; some from $LP^{color}_{i}$ and some from $LP^{depth}_{i}$. The contour cost $\triangle R_{ch_{i,1},ch_{i,2}}^{C}$ is computed by counting the number of contour elements (two neighboring pixels with different labels define one contour element between them) of the regions $ch_{i,1}$, $ch_{i,2}$ and multiplying for the average cost of encoding each contour element. Since we are considering the common contours, the contour cost for the parent is zero. Then:

\begin{equation} 
\triangle R_{ch_{i,1},ch_{i,2}}^{C} =  - c_{A}\,N_{ch_{i,1},ch_{i,2}} - c'_A\,N'_{ch_{i,1},ch_{i,2}}
\label{eq:RCi_cd}
\end{equation}
where $N_{ch_{i,1},ch_{i,2}}$ and $N'_{ch_{i,1},ch_{i,2}}$ are the numbers of common contours of the two children from the depth and color partitions respectively, while $c_A$ and $c'_A$ are the average costs to encode a contour element for either the depth and the color partitions. 
Note that contours from the color partition do not introduce any cost, therefore $c'_A=0$. The $c_A$ has been obtained experimentally using the multi-view sequences resulting in $c_A$ = 1.2 bits per contour position.

\begin{figure}[ht]
    \centering    \includegraphics[trim=0 30 0 100,clip,width=0.53\columnwidth]{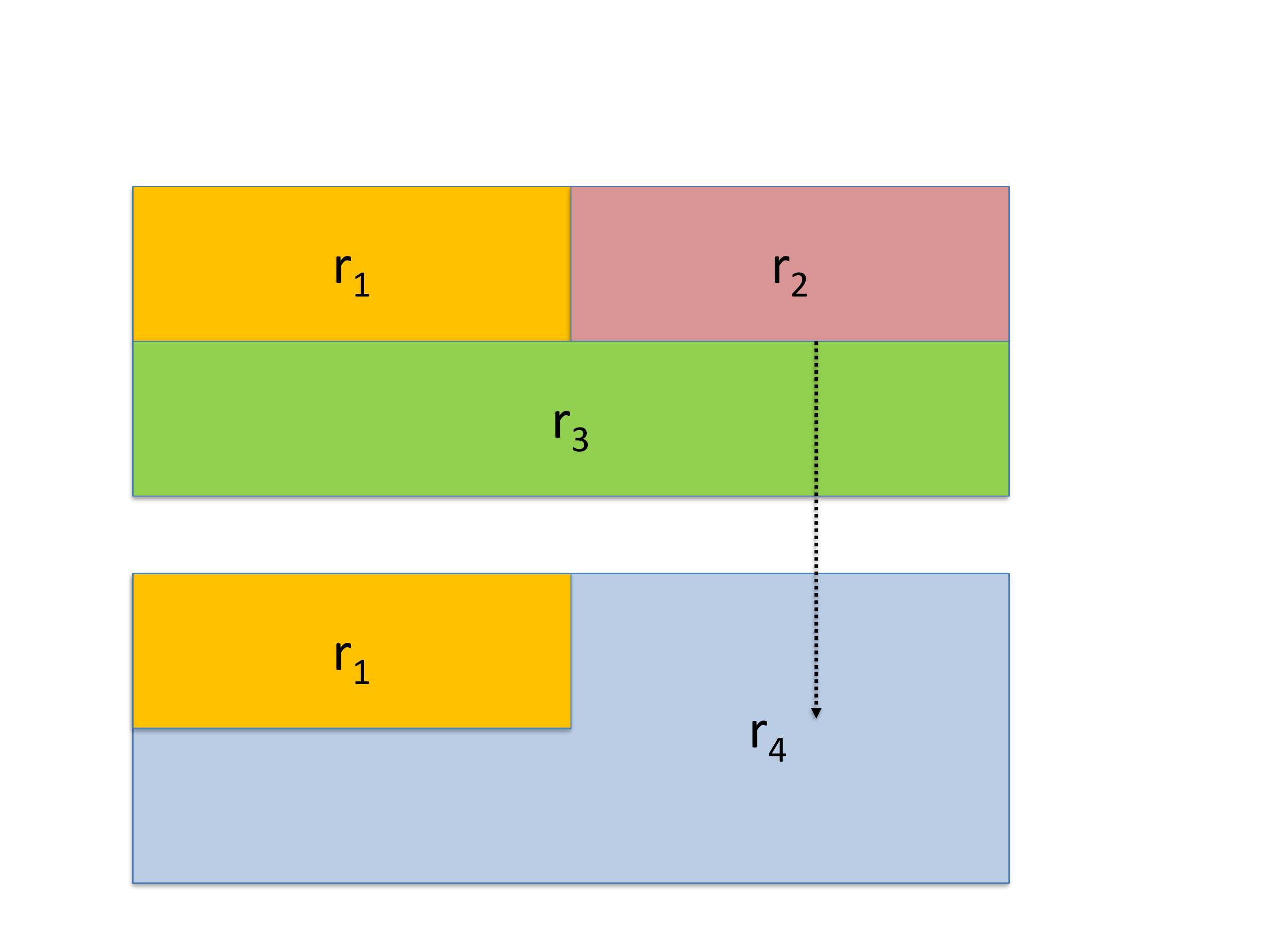}
    \caption[Multi-View representation: single-view rate]{The Single-View rate for each region is the rate saving promoted by the merging. Regions $r_2$ and $r_3$ are merged into $r_4$, leading to represent the depth map without the contour between $r_2$ and $r_3$ and saving one 3D plane.}
    \label{fig:1v_rate_planes}
\end{figure}

\equaref{eq:R} is computed as: 
\begin{equation} 
\triangle R_{ch_{i,1},ch_{i,2}}  =  \triangle R_{ch_{i,1},ch_{i,2}}^{C} + \triangle R_{ch_{i,1},ch_{i,2}}^{T} =  - c_{A}\,N_{ch_{i,1},ch_{i,2}} - R^{T}
\label{eq:Ri}
\end{equation}
\figref{fig:1v_rate_planes} shows an example of merging two regions. Encoding $r_4$ instead of $r_2$ and $r_3$ saves the cost of encoding the boundary between the regions and the texture coefficients of one region.

The color-depth optimization in each view addresses the addition of the depth boundaries in the color partition in an optimal fashion while reducing heavily the total number of regions. Doing this procedure in each view also reduces the number of occluded regions in the Multi-View Optimization. 

\subsection{Multi-View Optimization}
\label{sec:multi-view-opt}

\begin{figure}[ht]
    \centering    \includegraphics[trim=0 110 0 120,clip,width=\columnwidth]{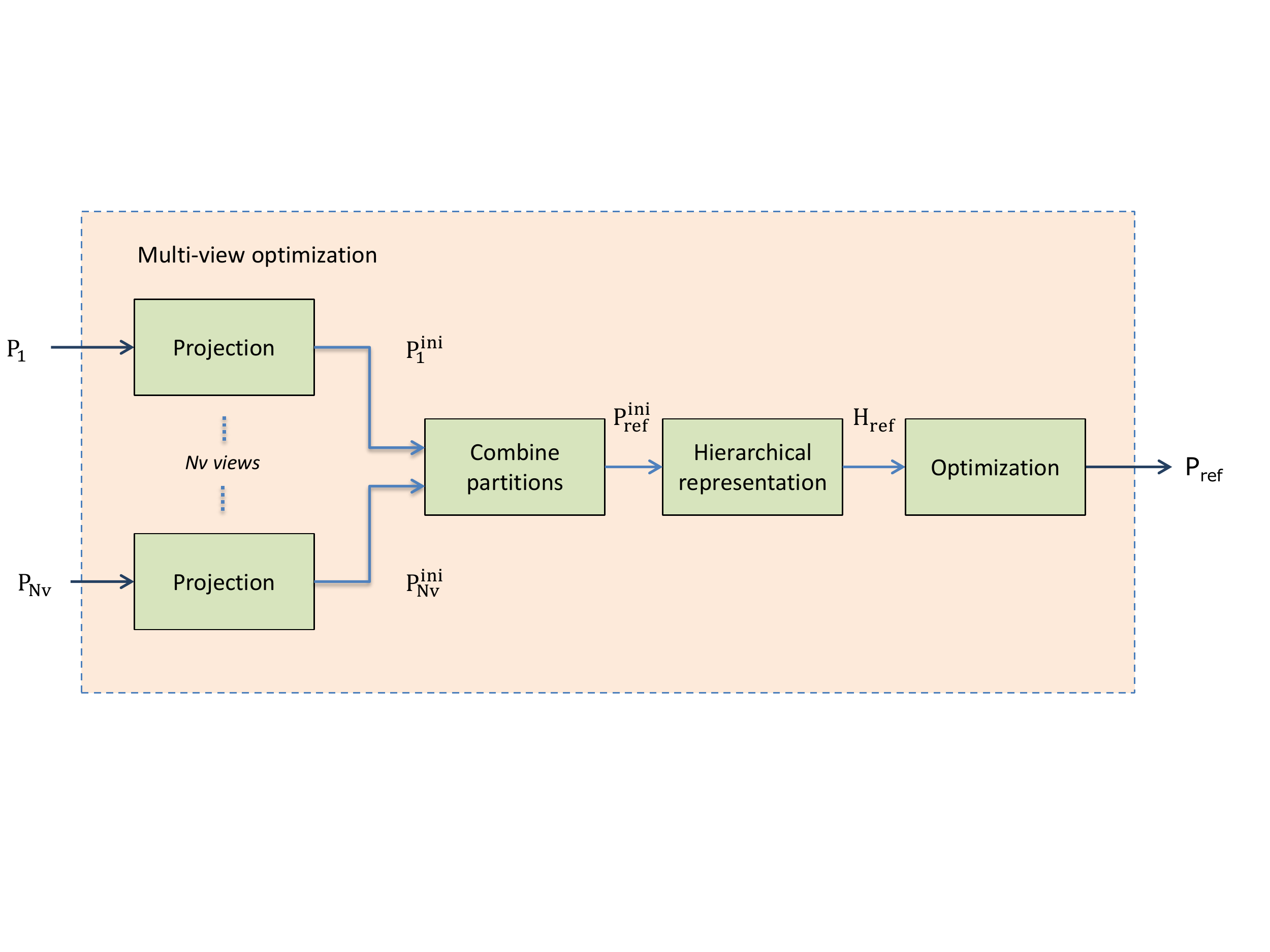}
    \caption[Multi-View representation: multi-view optimization]{Multi-View Optimization process for each view. Partitions from multiple views are combined in a unique hierarchy. Rate-Distortion measures are computed using all the input views and stored in the hierarchy to find the joint optimal partition for the multiple views.}
    \label{fig:multi-view-opt-scheme}
\end{figure}

The multi-view optimization process is shown in \figref{fig:multi-view-opt-scheme}. The partition $P_i$ of each of the $N_v$ views is projected, using the camera parameters, to $view_{ref}$, obtaining an initial partition $P_i^{\,ini}$ for each view $i$: 

\begin{equation} 
P_{i}^{\,ini} = Proj_{ref}(P_i)
\label{eq:pini_creation}
\end{equation}

\deleted[id=rm]{The projection step uses the camera parameters to translate the information of each view to the reference view $view_{ref}$.}
Pixels in each partition $P_{i}$ are processed in scan order and the corresponding pixel labels are projected to $view_{ref}$. Each individual projected partition $P^{ini}_i$ is defined by assigning to each pixel position the label of the nearest projected view $i$ pixel. With this process the regions that are near to the camera position prevail over the ones that are further away.

In this projection step, regions of view $i$ that are occluded in the reference view are detected. A region is considered occluded if the number of pixels of this region in $view_{ref}$ is less than half the number of pixels in view $i$. Occluded regions have no valid correspondence in $view_{ref}$. To solve this, these regions will be encoded independently in that view and removed from the projected partition in $view_{ref}$.

The individual projected partitions $P^{ini}_{i}$ in $view_{ref}$ are accumulated in a unique partition as stated in \equaref{eq:pini_ref}. 
\added[id=mm]{Unlike the single-view optimization, here in the projected $view_{ref}$ not all the pixels have a projected label. In order to obtain a partition with all the pixels assigned to a label, a hole filling algorithm creates new labels in-between regions with different label. For each combination of labels in each $view_{i}$, a new label is created in $P^{ini}_{ref}$.} 
\begin{equation} 
P^{ini}_{ref} = \bigcap_{i} P^{ini}_{i}
\label{eq:pini_ref}
\end{equation}

\begin{figure}[ht]
    \centering    \includegraphics[trim=0 148 0 20,clip,width=\columnwidth]{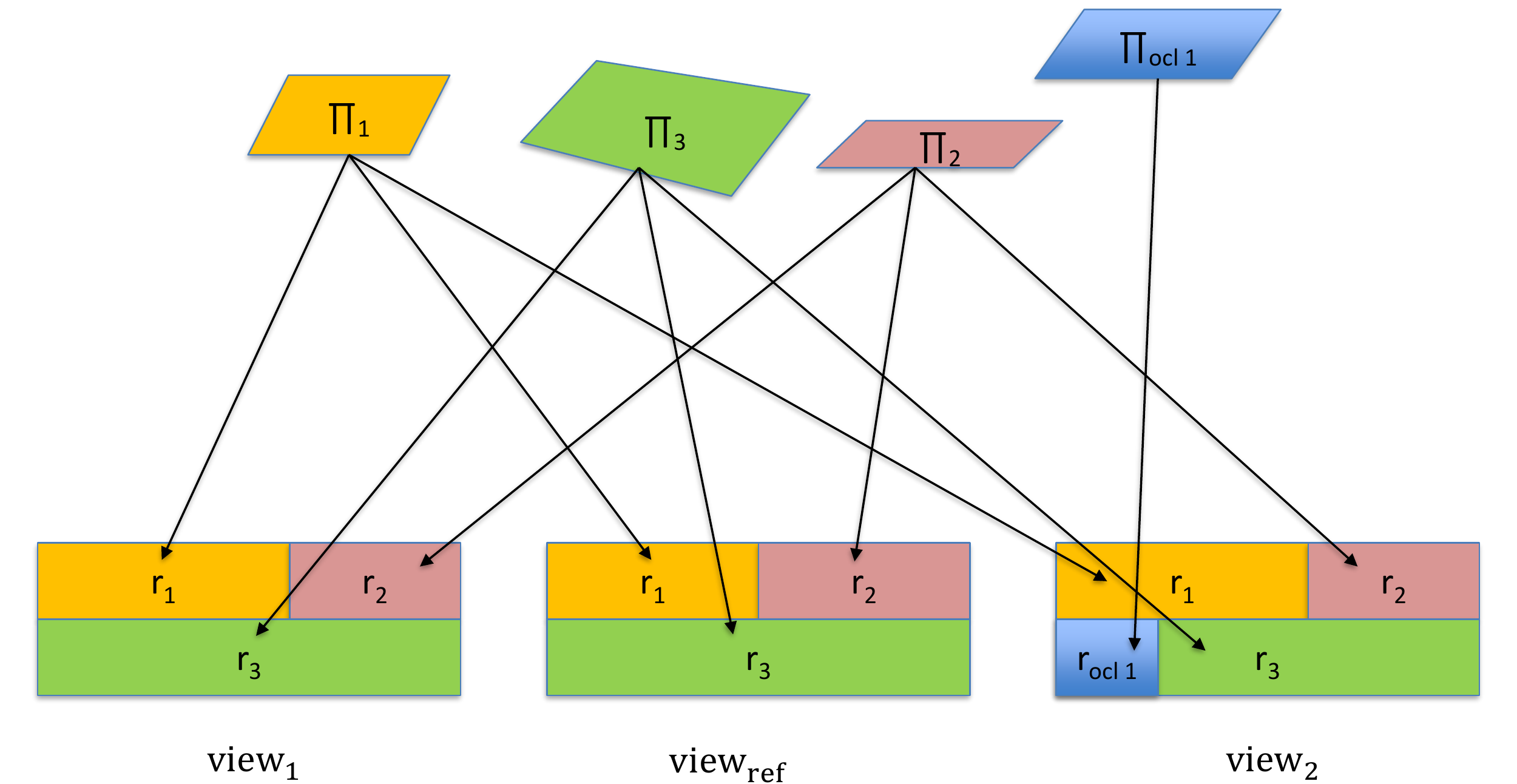}
    \caption[Multi-View representation: multi-view distortion]{The Multi-View distortion is computed projecting each 3D planar model to all the pixels of the corresponding region of each image.}
    \label{fig:mv_dist_planes}
\end{figure}

A hierarchical depth representation $H_{ref}$ is built using the accumulated partition $P^{ini}_{ref}$. Rate and distortion values for each region in $H_{ref}$ are found examining each partition generated in the $N$ views. The increment of distortion for each union is computed as: 

\begin{equation} 
\triangle D_{ch_{1},ch_{2}} = \sum\limits_{i=1}^{N} \triangle D_{ch_{i,1},ch_{i,2}} \label{eq:d_mv}
\end{equation}
where $\triangle D_{ch_{i,1},ch_{i,2}}$ represents the increment of distortion of each view $i$ as in \equaref{eq:increment_Di}. A graphical example of \equaref{eq:d_mv} is shown in \figref{fig:mv_dist_planes}. Notice that, in the example, $r_{ocl\,1}$ is occluded in $view_{ref}$. Therefore, $r_{ocl\,1}$ does not participate of the optimization process and is encoded with a independent plane.

Similarly, the rate is computed using the information of all views, adding  the boundary cost (from depth boundaries only, as $c'_A=0$ for color contours):

\begin{equation} 
\triangle R_{ch_{1},ch_{2}} = \sum\limits_{i=1}^{N} \triangle R_{ch_{i,1},ch_{i,2}} \label{eq:r_mv}
\end{equation}

\begin{figure}[ht]
    \centering    \includegraphics[trim=0 180 0 140,clip,width=0.8\columnwidth]{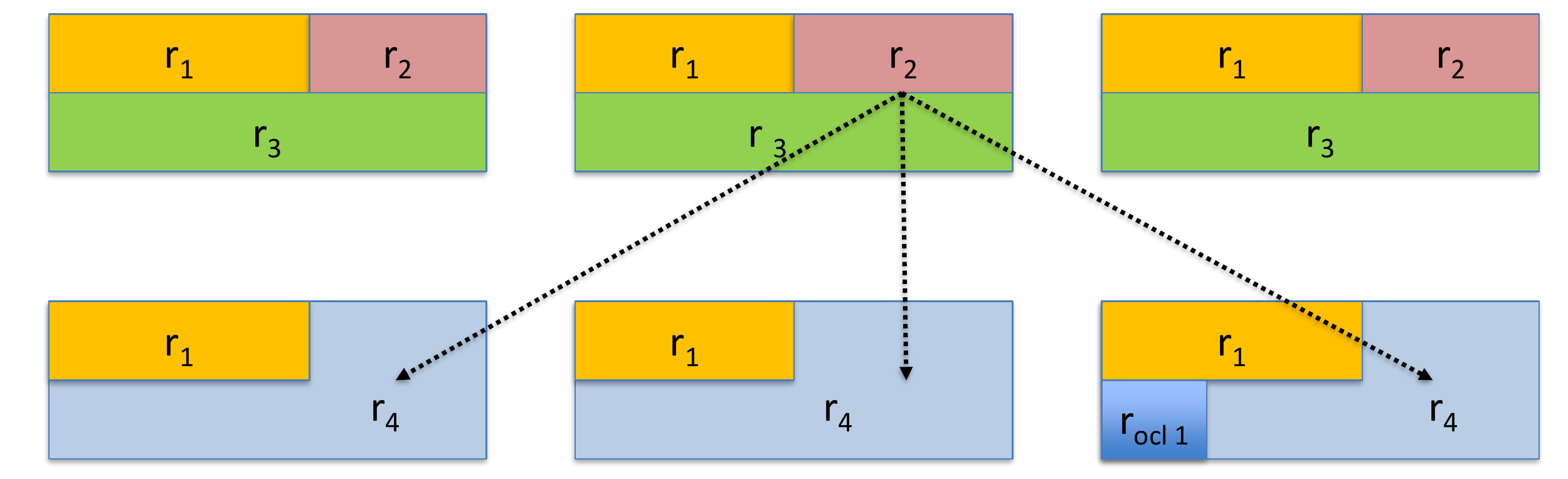}
    \caption[Multi-View representation: multi-view rate]{The Multi-View rate for each region is the rate saving obtained with the merging. A merging in $view_{ref}$ may force a merging in the other views.}
    \label{fig:mv_rate_planes}
\end{figure}

As the same region model encodes regions of multiple views, the texture rate is the same than in the single view optimization:

\begin{equation} 
\triangle R_{ch_{1},ch_{2}} = - c_{A}\,\sum\limits_{i=1}^{N} N_{ch_{i,1},ch_{i,2}} - R^{T}
\end{equation}
An example of computing the rate for the Multi-View optimization is shown in \figref{fig:mv_rate_planes}.

The $P_{ref}$ partition is obtained using the hierarchical optimization of Section~\ref{sec:hierarchy_rd}:

\begin{equation} 
P_{ref} = Opt_{\lambda}(H_{ref})  \label{eq:opt_mv_eq}
\end{equation}


\replaced[id=mm]{Since the hierarchy $H_{ref}$ is build in the $view_{ref}$, the merging orders are determined in $view_{ref}$. By incorporating the information of the $N_v$ views to the optimization process, the resulting partition $P_{ref}$ defines an optimal partition in all the views.}{Since the hierarchy $H_{ref}$ is built in the $view_{ref}$, the resulting partition of the optimization process $P_{ref}$ defines a partition in $view_{ref}$. Notice that this information is reproduced to the $N_v$ views by replicating the information about active contours in $view_{ref}$ to the $N_v$ views. As the hierarchy in $view_{ref}$ incorporates the rate and distortion measures of the multiple views, the resulting $P_{ref}$ partition projected to the different viewpoints defines an optimal representation to the $N_v$ views. }

\replaced[id=mm]{The optimization parameter $\lambda$ that defines the Rate-Distortion trade-off is the quality parameter. By varying $\lambda$, different Rate-Distortion points are found, being able to obtain 3D scene representations with different level of detail. }{The optimization parameter $\lambda$ that defines the Rate-Distortion trade-off is our quality parameter. By varying $\lambda$, different Rate-Distortion points are found, being able to obtain 3D scene representation with increasing quality.} 
\deleted[id=rm]{An overview of the partitions obtained at every stage is shown in \figref{fig:fancy_scheme}.}

\section{Multi-view depth map coding}
\label{sec:multi-view-coding}

In Section~\ref{sec:mv3Drepres}, the proposed method to obtain a joint segmentation of the scene combining color and depth partition has been explained. In order to convey the depth information to the decoder we propose a method to pack the information of the multiple views efficiently. \added[id=mm]{The method assumes that the camera parameters and the color images of each view have been encoded and sent to the decoder before decoding the depth map information. 
The partition $P_{ref}$ is projected back, following the inverse process of \equaref{eq:pini_creation}, to obtain a consistent set of coding partitions $P^{cod}_i$ for each of the $N_v$ views.}


\begin{figure}[htb]
\centering
\includegraphics[trim = 0mm 40mm 30mm 60mm, clip,width=0.95\textwidth]{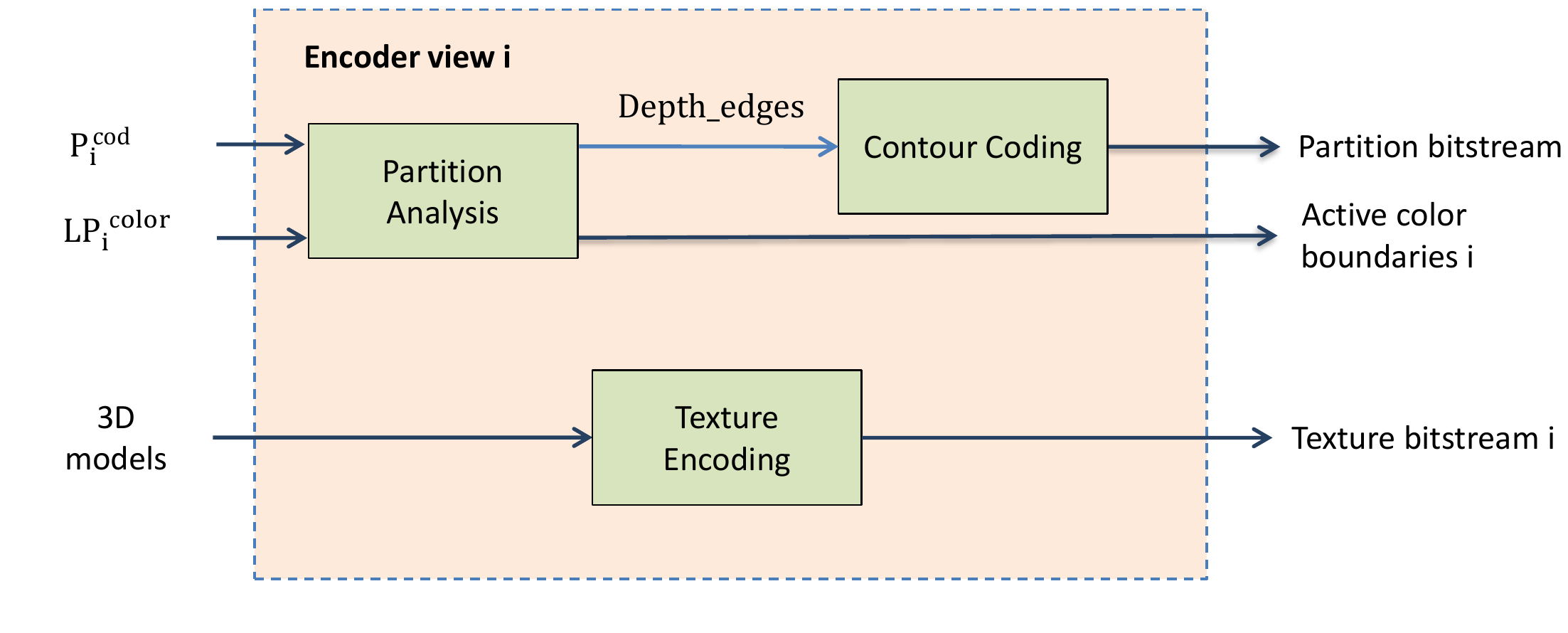}
\caption[Multi-View coding: encoder scheme]{Encoder scheme. Partition information is generated independently for each view. Texture encoding encodes the 3D plane in INTRA if it is the first view where it appears or INTER if it has been encoded previously.}
\label{fig:encoder_view}
\end{figure}

\begin{figure}[ht]
    \centering    \includegraphics[trim=0 150 0 100,clip,width=1\columnwidth]{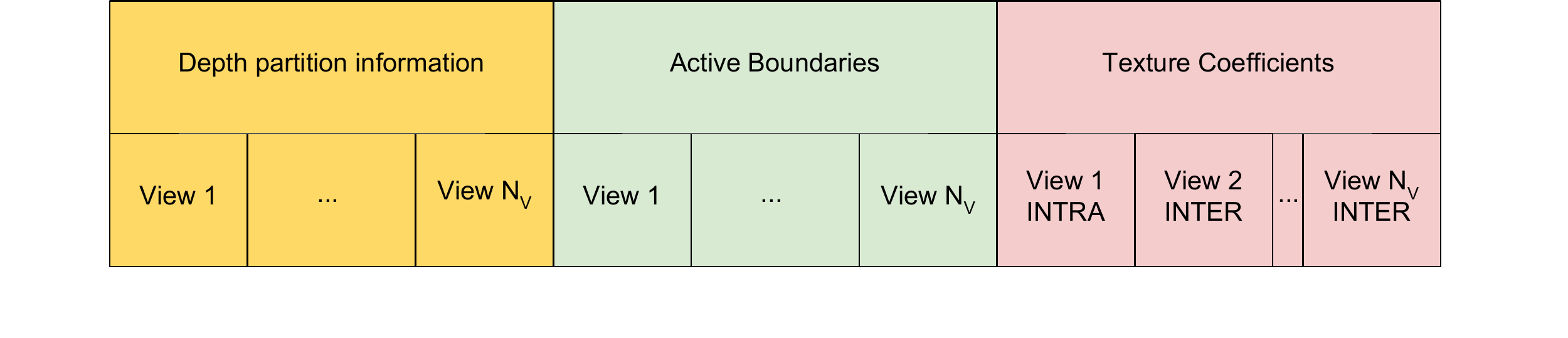}
    \caption{Bitstream with the information of the $N_v$ views. The depth partition information signal the depth boundaries added to the color information. With the active boundaries information the final coding partition is obtained. The texture coefficients for views 2 to $N_v$ make use of the previous transmitted coefficients.}
    \label{fig:bitstream}
\end{figure}

\replaced[id=mm]{The general coding scheme for each view $i$ is presented in \figref{fig:encoder_view}. The first step consists in analyzing which contours from $LP^{color}_{i}$ and $LP^{depth}_{i}$ are present in $P^{cod}_i$. Contours in $P^{cod}_i$ from $LP^{depth}_{i}$ are encoded with Freeman contour coding~\cite{Freeman1961} independently for each view. For the color contours, information of color boundaries that are not active (because of region mergings) is sent to the decoder indicating if the merging has been done or not. Active information is stored with 1 bit per each merging in $H_{ref}$.}{To generate the partition, the first step consists in signaling the depth boundaries (depth partition information in \figref{fig:bitstream}). The information of the depth boundaries added in the multi-view optimization is encoded with Freeman contour coding~\cite{Freeman1961}. The contours for each view are encoded independently.}

\deleted[id=rm]{The final coding partition for each view do not include all the color boundaries. 
To remove these boundaries, 
the information of active boundaries for each view is generated and sent to the 
decoder. This information consist in a binary variable for each merging in the hierarchy, indicating if the union is active or not.}

Once the partition information has been encoded, the texture coefficients are generated. The first view encodes the texture information of all the planes associated to that view in INTRA mode, while planes from the subsequent views will be encoded using INTER view prediction when possible. In the INTRA mode, the 3D planes are represented by using the distance of that plane to a camera plus the plane orientation. This plane orientation is stored in spherical coordinates with their 3D angles $\theta$ and $\phi$: \\
\begin{equation} 
\theta = arccos \left(\frac{n_z}{\|n\|}\right) \qquad
\phi = arctan \left(\frac{n_y}{n_x}\right)
\end{equation}

where the $n_z$ component is pointed towards $z > 0 $. The resulting angles (in the range $ 0 \le \theta \le \frac{\pi}{2} $ and $ 0 \le \phi \le 2 \pi $) are encoded with equal precision with a uniform quantizer.

The distance from the plane to the camera is converted to an alternative quantized representation using the distance to depth map conversion:
\begin{equation} 
 C_{dist} = \frac{1.0}{\frac{d(pl,c)}{(2^{N_{dist}}-1)}*(\frac{1.0}{MinZ} - \frac{1.0}{MaxZ}) + \frac{1.0}{MaxZ}} 
\end{equation}

where $d(pl,c)$ is the euclidean distance between the region plane and the camera, $N_{dist}$ is the number of bits to be used in the quantization and $MaxZ$ and $MinZ$ are the maximum and minimum depth values of the image. 

For subsequent views $i$, the INTER mode is used. Planes from the first view are projected to view $i$ as done in the construction of the multi-view partition in $view_{ref}$. Each region in view $i$ has a candidate plane from the first view. This plane is compared with a plane coded with INTRA in view $i$. The plane with the lower error is kept. If the plane is coded with INTER, a SKIP mode is sent, if not the region is INTRA coded.

\added[id=mm]{The bitstream (depicted in \figref{fig:bitstream}) provides the decoder with the information needed to recover the partition and the texture for each region. The partition bitstream, active color boundaries and texture bitstream for each view are provided in addition of the number of regions $N_{regs\_color}$ in each initial color partition $LP^{color}_{i}$. $N_{regs\_color}$ is the same for the $N_v$ views.}

\begin{figure}[htb]
\centering
\includegraphics[trim = 0mm 40mm 0mm 50mm, clip,width=0.95\textwidth]{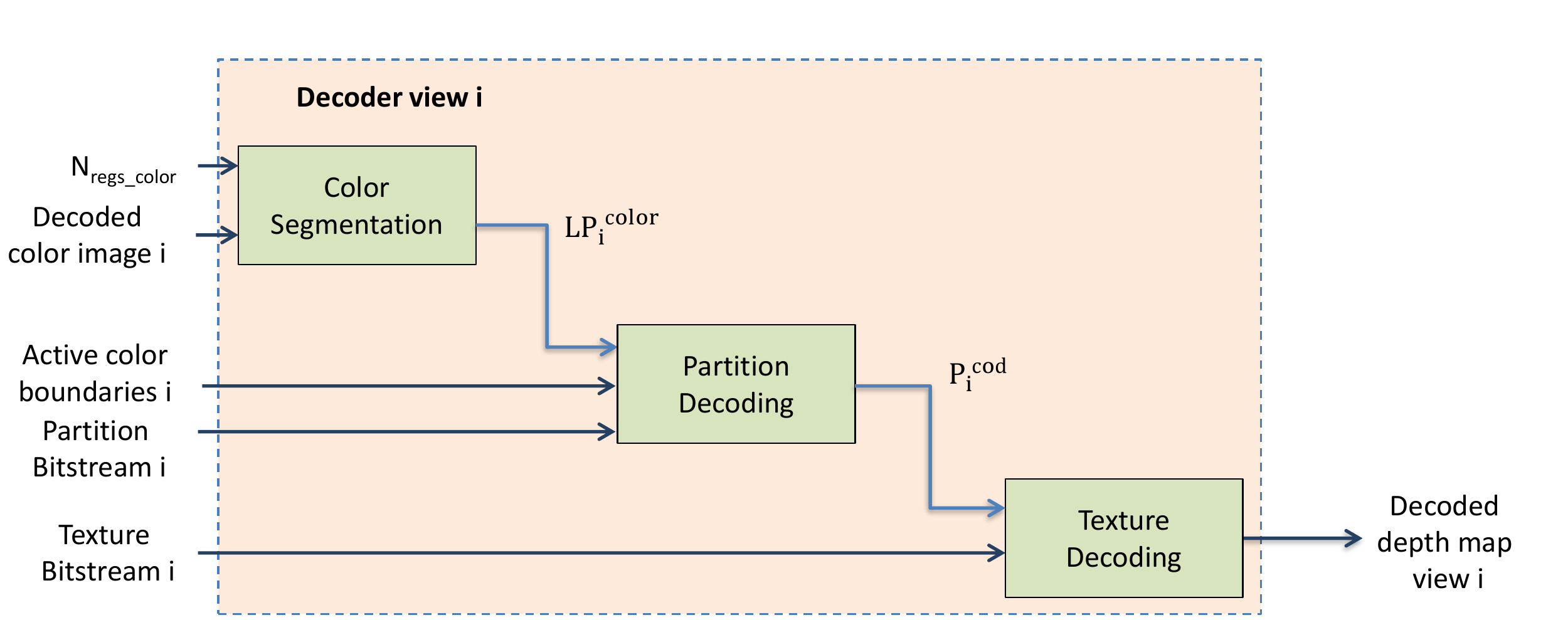}
\caption[Multi-View coding: decoder scheme]{Decoder scheme. $LP^{color}_{i}$ is generated from the decoded color image using $N_{regs\_color}$. Using the information of active color boundaries and the partition bitstream the encoding partition $P^{cod}_i$ is generated. The decoded depth map for view i obtained by projecting the 3D planes to $P^{cod}_i$.}
\label{fig:decoder_mv}
\end{figure}

\added[id=mm]{The decoder process is depicted in \figref{fig:decoder_mv}. For each view, the $LP^{color}_{i}$ is built from the decoded color image. From that partition, color edges are removed using the active color boundaries information. Then new edges are added from the partition bitstream. Depth map is recovered by decoding the texture bitstream and projecting each 3D plane to $P^{cod}_i$.}

\section{Experimental Evaluation}\label{sec:results}

In this section, we present a quantitative evaluation of our multi-view optimization segmentation and coding method. We use the RGB-D multi-view sequences
\emph{Ballet}, \emph{Undo Dancer}, \emph{Breakdancers} and \emph{Ghost Town} for
evaluation~\cite{Zitnick2004,zhang2011ghost,rusanovskyy2011undo}. 
Each sequence is composed of color and depth frames from up to $8$ camera views as
well as camera parameters for each frame. \emph{Ballet} and \emph{Breakdances} have been captured using color (RGB) cameras and the depth has been computed afterwards. \emph{Undo Dancer}
and \emph{Ghost Town} are synthetic sequences. The resolution of the sequences is 1920x1088 pixels for \emph{Undo Dancer} and~\emph{Ghost Town} and 1024x768 pixels for
\emph{Ballet} and \emph{Breakdancers}. We assume that color views have been previously decoded at the receiver side. The different stages of the proposed method are
evaluated using $25$ frames of each sequence.

In order to assess our technique, we first explore different configurations of the proposed method in Section~\ref{sec:evaluation-config}. Finally, our proposal is compared against \emph{HEVC}, \emph{3D-HEVC} and \emph{MV-HEVC} in Section~\ref{sec:evaluation-rd}.

Experiments will make use of three views: $view_{left}$, $view_{right}$ and $view_{ref}$ using the same configuration of \emph{3D-HEVC}. $view_{left}$ and $view_{right}$ are the two base views to encode and $view_{ref}$ is the location selected to perform the optimization process in the multi-view optimization block (See Section~\ref{sec:multi-view-opt}). This view corresponds to the virtual view of \emph{3D-HEVC}. Color and depth map images as well as the camera parameters are available in the three views considered.

\subsection{Configuration}
\label{sec:evaluation-config}

In this Section, the different blocks of the overall system are analyzed. The first
        experiment studies how the number of views affects the coding performance of our method. 
        Then, we show the gain of using the RD optimization in the hierarchy. In the third experiment, we
        show the difference between using a color partition or a combination of color
        plus depth partitions. Finally we study the distribution of rate among depth partition
        information, active boundaries and texture coefficients. Results for the different sequences are
        averaged into a single Rate-Distortion curve. 



\subsubsection*{Combining multiple views}

\replaced[id=rm]{In this experiment we study the combination of several views prior to the multi-view optimization}{In this experiment we study the use of a single-view optimization prior to the multi-view optimization.}

\begin{figure}[h!]
    \centering    \includegraphics[width=0.99\columnwidth]{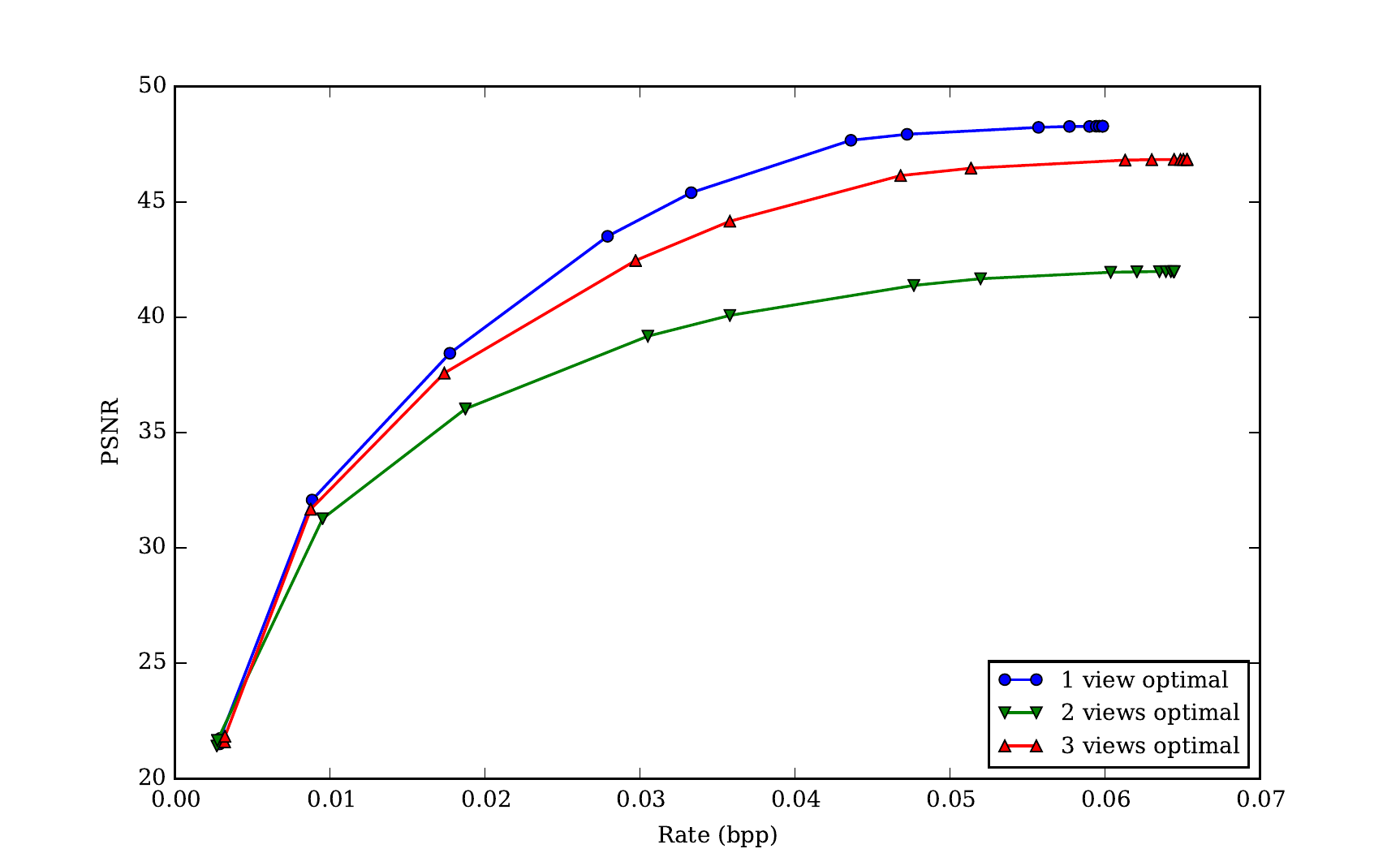}
    \caption{Coding results in the $view_{ref}$. 1 view configuration has only information of $view_{ref}$, 2 view configuration use the 2 views projected to $view_{ref}$ and 3 view configuration uses the three of them.}
    \label{fig:1view_vs_mv_depth_vv}
\end{figure}

\replaced[id=rm]{\figref{fig:1view_vs_mv_depth_vv} shows the results of encoding the $view_{ref}$ depth map in different configurations. In each configuration, a different number of views is used to obtain the hierarchy in $view_{ref}$.}{Firstly we show the results of building a hierarchy in the $view_{ref}$ for different configurations. Then this information is projected back to $view_{left}$ and $view_{right}$.
\figref{fig:1view_vs_mv_depth_vv} shows the results of combining the information of a different number of views into $view_{ref}$.}

\deleted[id=rm]{Initial depth partitions are used in the 3 views.}
In the \emph{1 view optimal} configuration, only information from the $view_{ref}$ is used. 
This configuration is the one that achieves better Rate-Distortion figures. 
The \emph{2 views optimal} configuration uses information of $view_{left}$ and $view_{right}$. As the projection from those views to $view_{ref}$ is not able to match all the depth boundaries in $view_{ref}$, the results for the \emph{2 views optimal} are below the 1 view configuration. With the \emph{3 views optimal} all 3 views are used. Results are slightly below the \emph{1 view optimal}, but in this case we optimize for the 3 views together, obtaining a consistent segmentation across the views.

\begin{figure}[h!]
    \centering    \includegraphics[width=0.99\columnwidth]{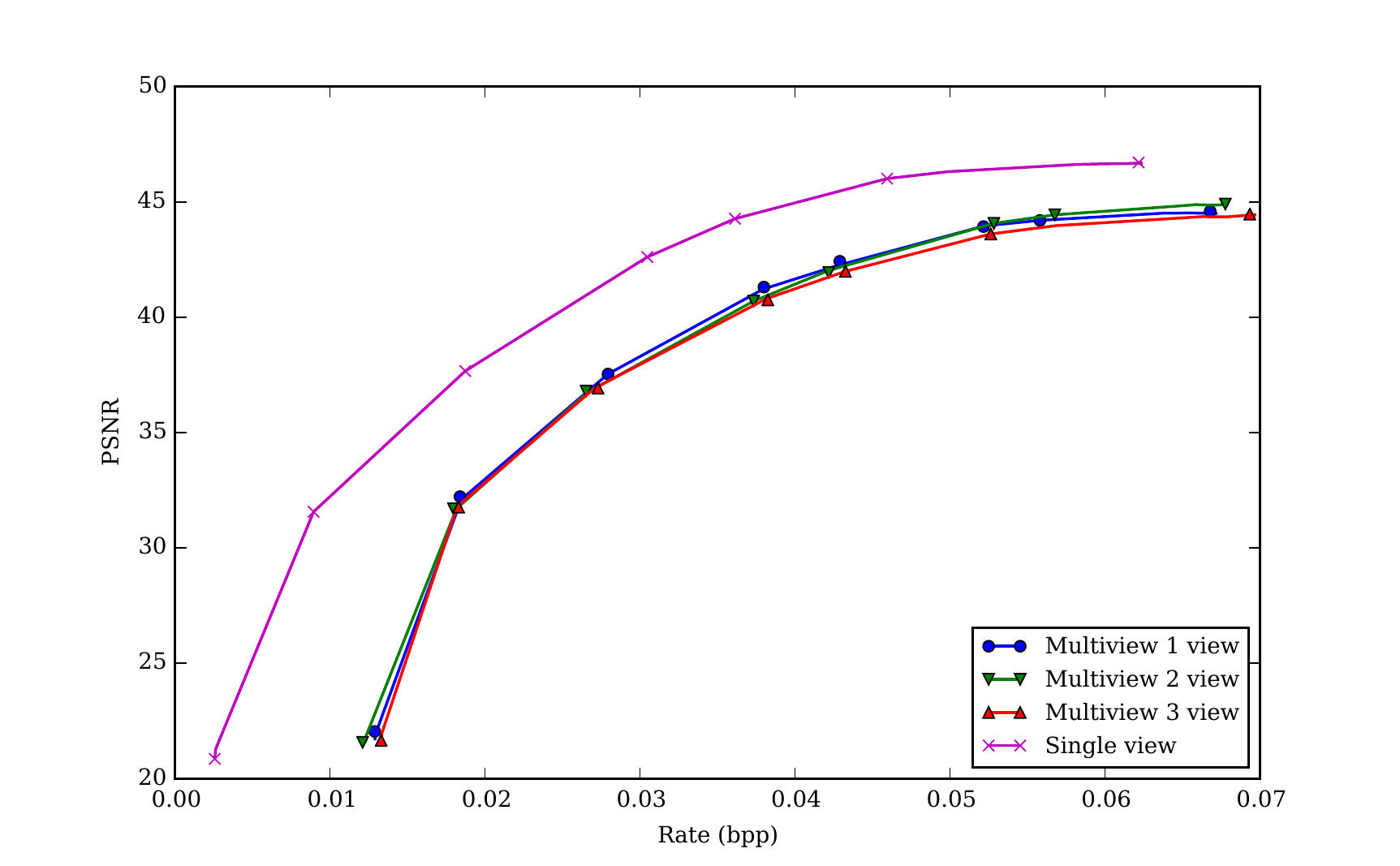}
    \caption{Results from \figref{fig:1view_vs_mv_depth_vv} are translated to $view_{left}$ and $view_{right}$. The three multi-view configuration (all with hierarchy built in $view_{ref}$) are compared with a hierarchy in each view independently.}
    \label{fig:1view_vs_mv_depth_side}
\end{figure}

\replaced[id=rm]{The resulting optimal coding partitions for each configuration are projected back to $view_{left}$ and $view_{right}$, as presented in Section~\ref{sec:multi-view-coding}.}{The resulting partitions in \figref{fig:1view_vs_mv_depth_vv} are projected back to $view_{left}$ and $view_{right}$ as presented in Section~\ref{sec:multi-view-coding}.}
Results are shown in \figref{fig:1view_vs_mv_depth_side}. 
The different configurations achieve similar results, all of them below the results of the single-view configuration. This behavior is due to the fact that the hierarchical optimization in $view_{ref}$ restricts the possible mergings in the other views and that an extra rate is needed to encode occluded regions.

With the Single-View optimization, the number of regions in the different views is reduced before building the shared hierarchy between views in $view_{ref}$. Being able to merge regions in a single-view reduces the number of regions that are occluded in $view_{ref}$ while creating a hierarchy $H_{ref}$ in $view_{ref}$ with relevant regions.

\subsubsection*{Rate-Distortion optimization}
\begin{figure}[ht]
    \centering    \includegraphics[width=0.99\columnwidth]{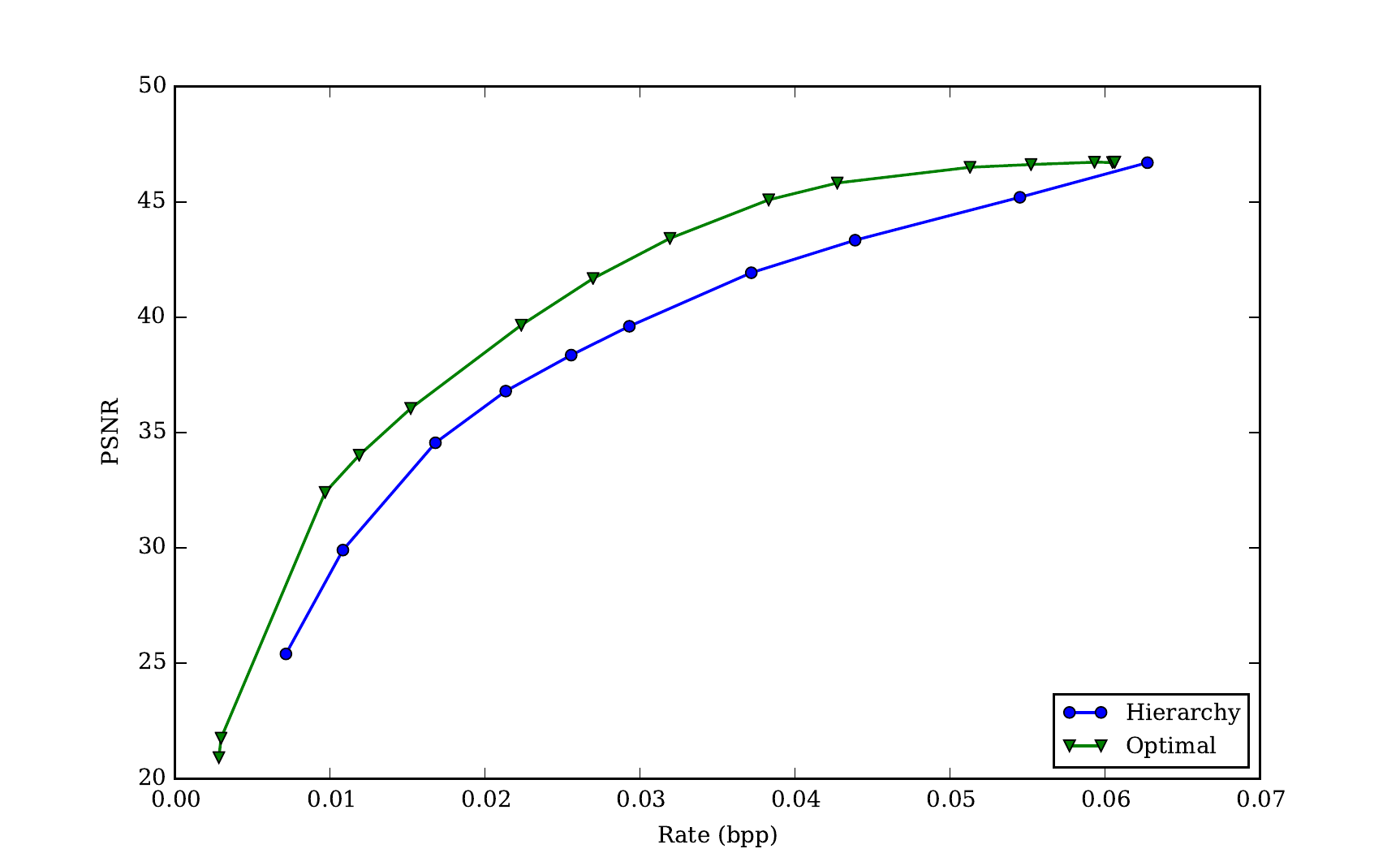}
 \caption{Rate-Distortion results with the proposed optimization method. Using an initial $H^{depth}$, results from coding the partition with 3D planes following the merging sequence and with the proposed optimization.}
\label{fig:merg_seq_vs_opt_depth}
\end{figure}

\replaced[id=mm]{In this Section, the performance of the Rate-Distortion optimization method in the Single-View Configuration is studied. Different cuts of the hierarchy $H^{c\_d}_{i}$ \added[id=rm]{(this is, different partitions)} extracted following the merging sequence are compared with optimal partitions $P_{i}$ with different quality parameter $\lambda$. Results are shown in \figref{fig:merg_seq_vs_opt_depth}. By using the proposed optimization, we are able to achieve a gain of 2-3 dB over using the $H_i^{depth}$ merging sequence.}{In this Section, the performance of the Rate-Distortion optimization method in the Single-View Configuration is studied. Different cuts of the hierarchy $H^{c\_d}_{i}$ are extracted following the merging sequence are compared with optimal partitions $P_{i}$ with different quality parameter $\lambda$. By using the proposed optimization, we are able to achieve a gain of 2-3 dB over using the $H_i^{depth}$ merging sequence.}


\subsubsection*{Color and depth combination}


\begin{figure}[ht]
    \centering    \includegraphics[width=0.99\columnwidth]{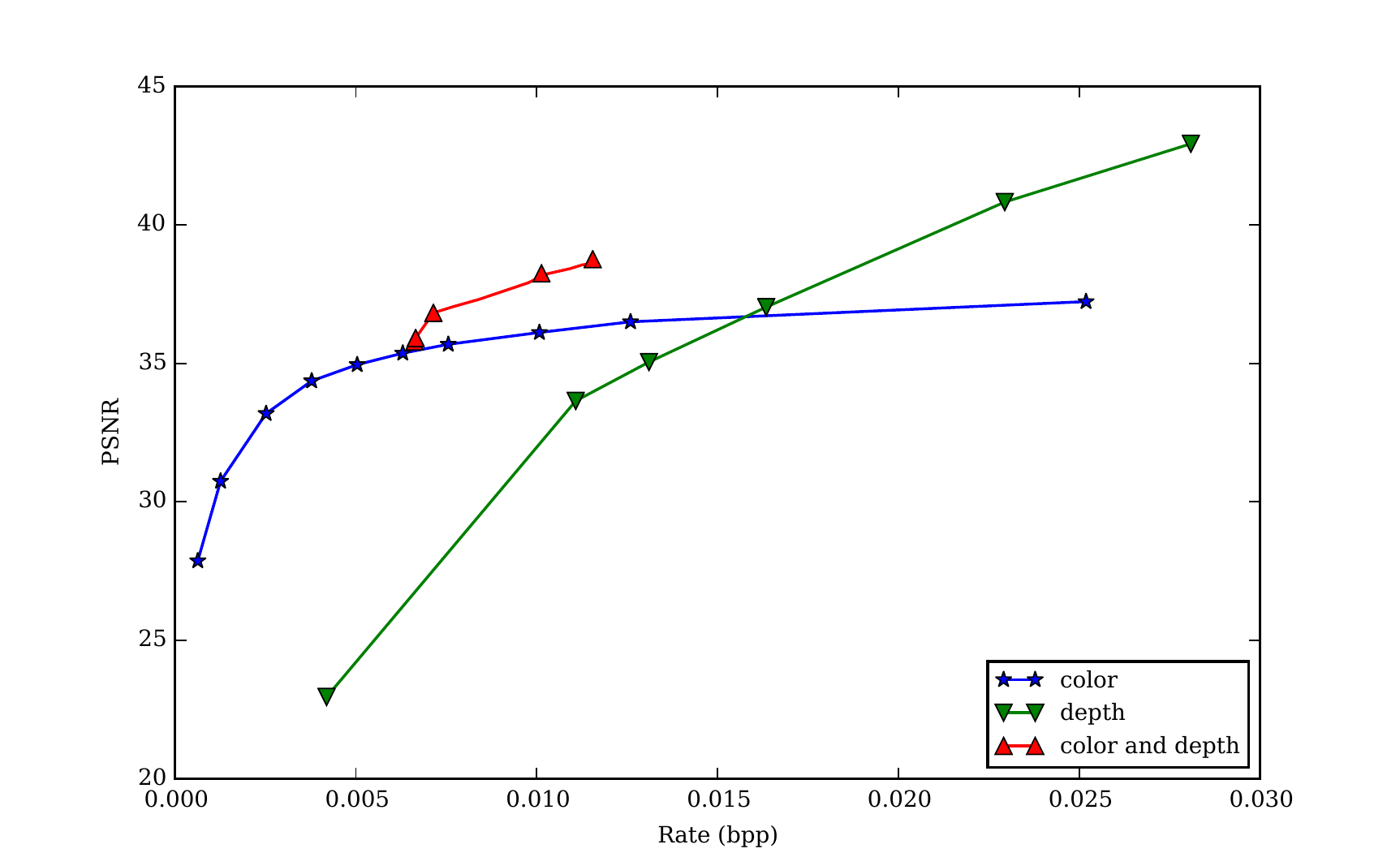}
    \caption{R-D curves using \textit{color} partitions, \textit{depth} partitions and combined \textit{color and depth} partitions to encode depth maps. Partitions using depth include the cost of encoding texture and boundary while the colors only need texture information.}
    \label{fig:color-depth-costs}
\end{figure}

\figref{fig:color-depth-costs} shows the R-D curves for three different options: using only a depth partition, using only a color partition and using a color-depth partition. Using only color partition has the advantage that no contour information has to be sent. It's a good option for very low bit-rates but the lack of precision in the region boundaries limits the maximum achievable quality. Using only depth information allows to obtain higher quality as all the contours have to be sent, the curve is displaced to the right. The combination of color and depth provides a good trade-off because most of the depth contours can be approximated using color boundaries and only a reduced set of depth contours not present in the color partition must be sent.

\subsubsection*{Multi-view optimization: rate statistics}

\begin{figure}[ht]
    \centering    \includegraphics[trim=0 0 80 0,clip,width=0.85\columnwidth]{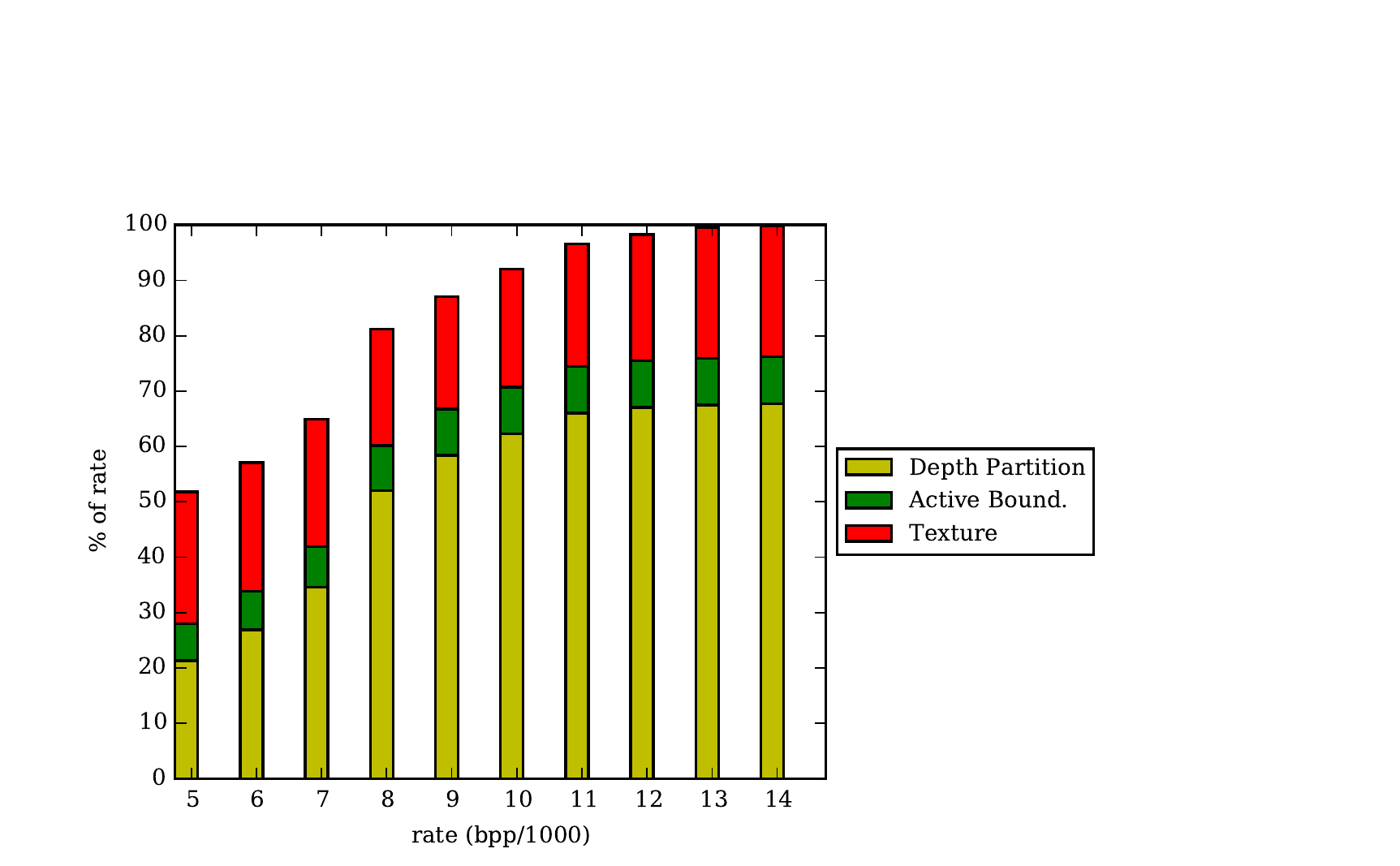}
    \caption{Distribution of rate in texture, partition and active boundaries information for different Rate-Distortion points.}
    \label{fig:rate-stats}
\end{figure}

To evaluate the distribution of rate among depth partition information, active boundaries and texture coefficients, the bitstream presented in Section~\ref{sec:multi-view-coding} was analyzed to determine the different contributions. \figref{fig:rate-stats} shows the rate distribution among each category. Each bar is normalized with the maximum rate in the last column. The cost of encoding the texture and the information of active boundaries remains fairly constant among all the different Rate-Distortion points. However, the partition cost becomes the prevailing one as the rate increases. The results of the optimization depend on the given budget rate. When the budget is low, only color boundaries are added. When more bits can be allocated, the optimization adds the costly depth boundaries.

\subsection{Rate-Distortion results}
\label{sec:evaluation-rd}

As the proposed method does not have temporal prediction, only intra modes for the different methods are taken. For each sequence, three views are used, $view_{left}$ and $view_{right}$ are encoded (each view independently) and the middle one is employed as the location for the $view_{ref}$. The view rendered using the original depth maps serves as a reference for the different methods in the virtual view.

To objectively evaluate the proposed method, error measures are taken both in the depth map and in the synthesized virtual view. Since depth maps are not visualized but used to render new images, the measure in the virtual view gives the real performance of the method. Measuring directly on the depth map gives an overview of how the original depth map can be represented with planes. Similar results were obtained using \emph{PSNR} and \emph{SSIM}. As both measures show similar tendencies, only the \emph{PSNR} will be shown to evaluate the error.

\begin{figure*}[!htbp]
  \begin{center}
    \begin{tabular}{cc}
      		\includegraphics[trim=35 0 35 0,clip,width=0.46\textwidth]{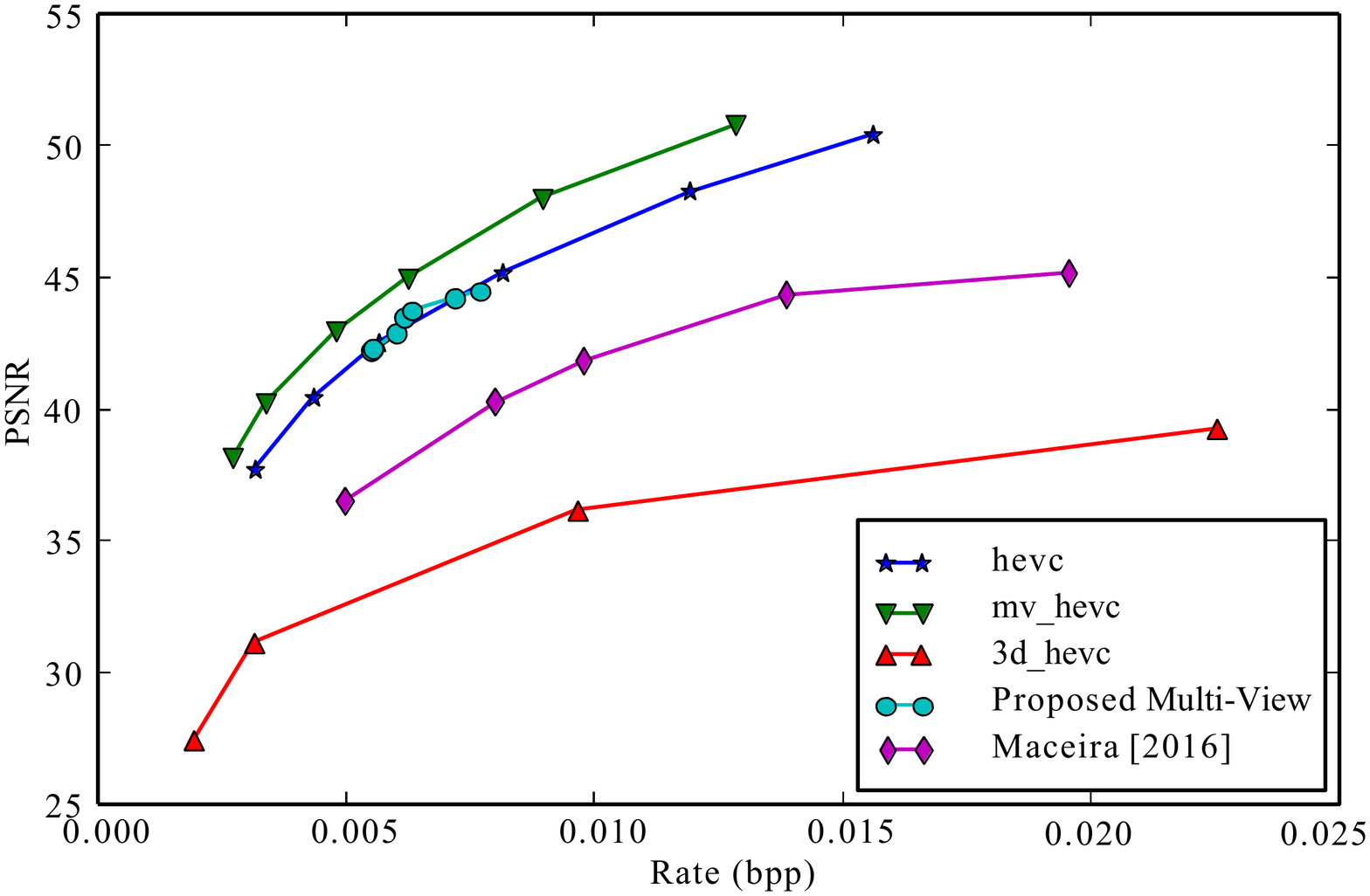}&
      		\includegraphics[trim=35 0 35 0,clip,width=0.46\textwidth]{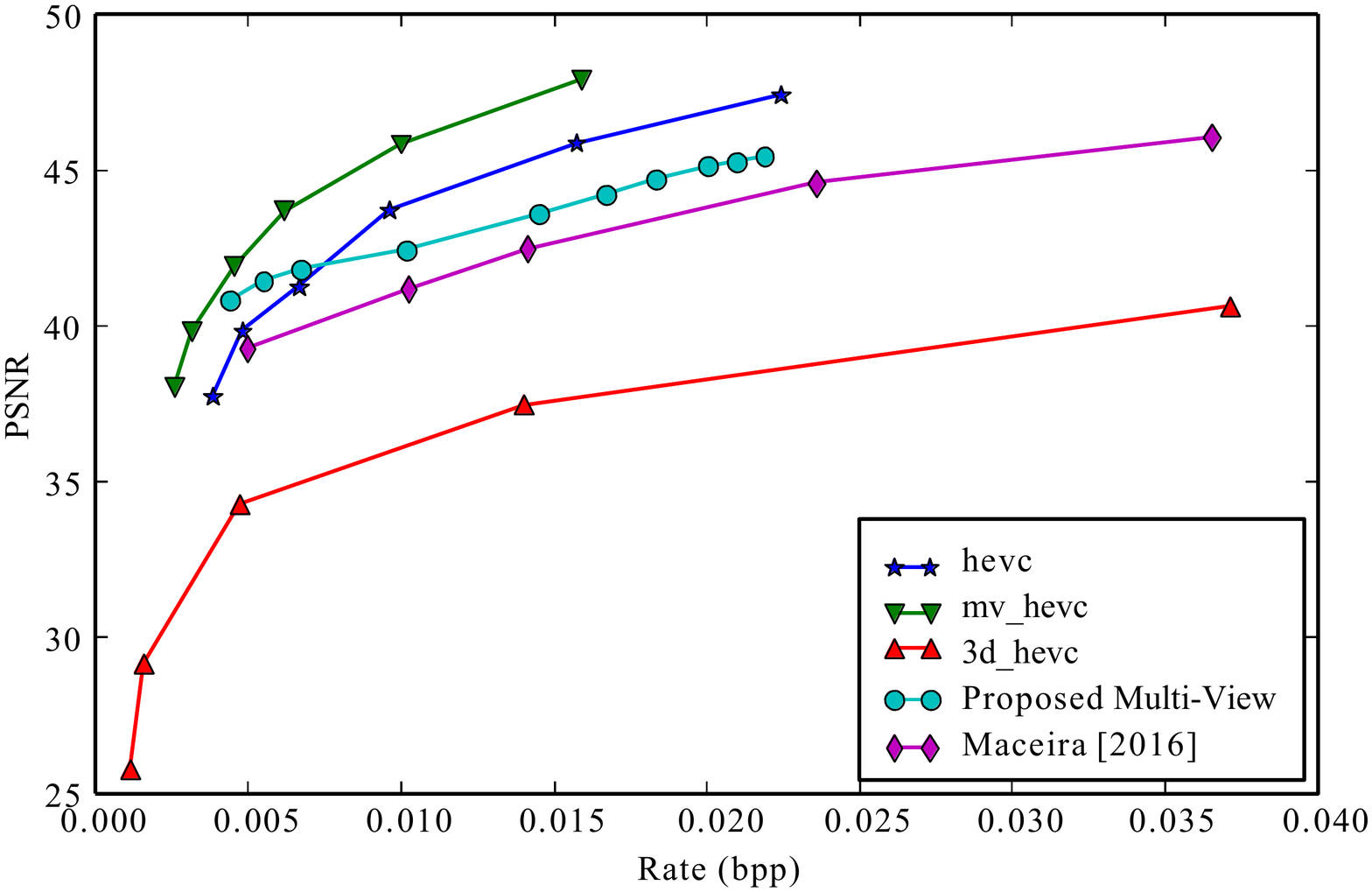}\\
	     	a) \emph{Undo Dancer} & b) \emph{Ghost Town}\\
      		\includegraphics[trim=35 0 35 0,clip,width=0.46\textwidth]{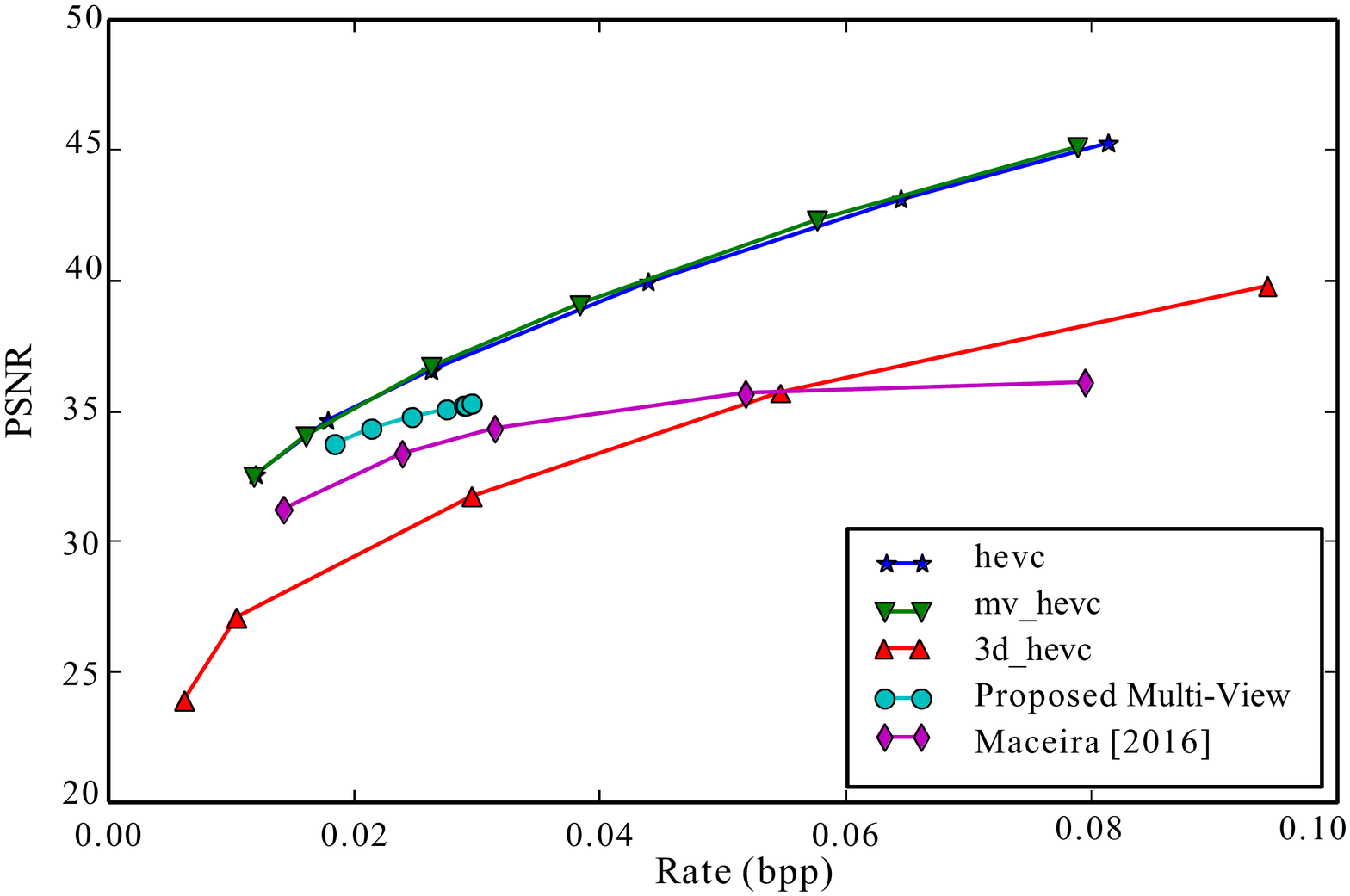}&   
      		\includegraphics[trim=35 0 35 0,clip,width=0.46\textwidth]{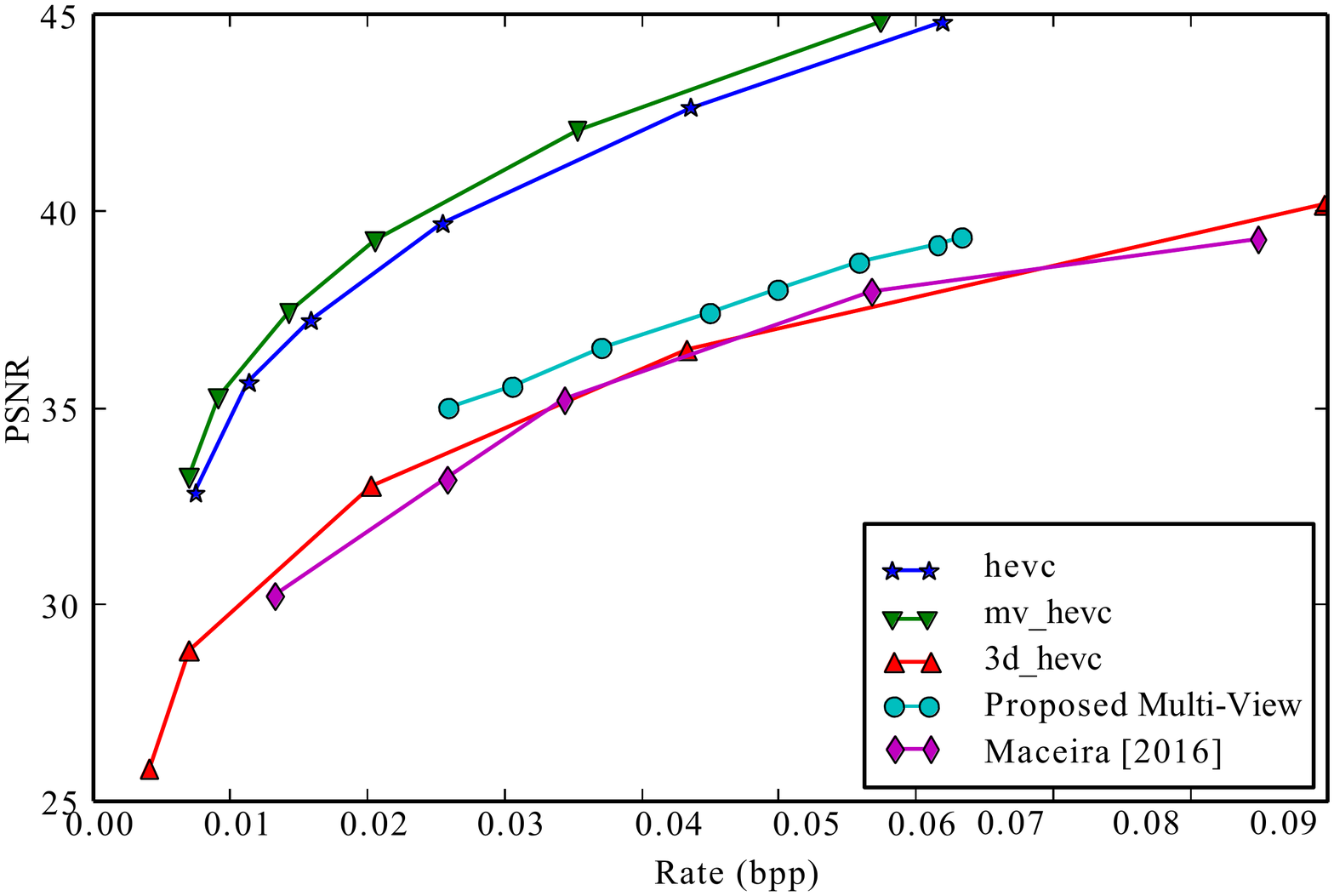}\\
	   	c) \emph{Ballet} & d) \emph{Breakdancers}\\
    \end{tabular}  
  \end{center}
  \caption{Rate distortion results evaluated directly in depth maps.}
  \label{fig:rate_dist_depth_map}
\end{figure*}

\begin{figure*}[!htbp]
  \begin{center}
    \begin{tabular}{cc}
      		\includegraphics[trim=35 0 35 0,clip,width=0.46\textwidth]{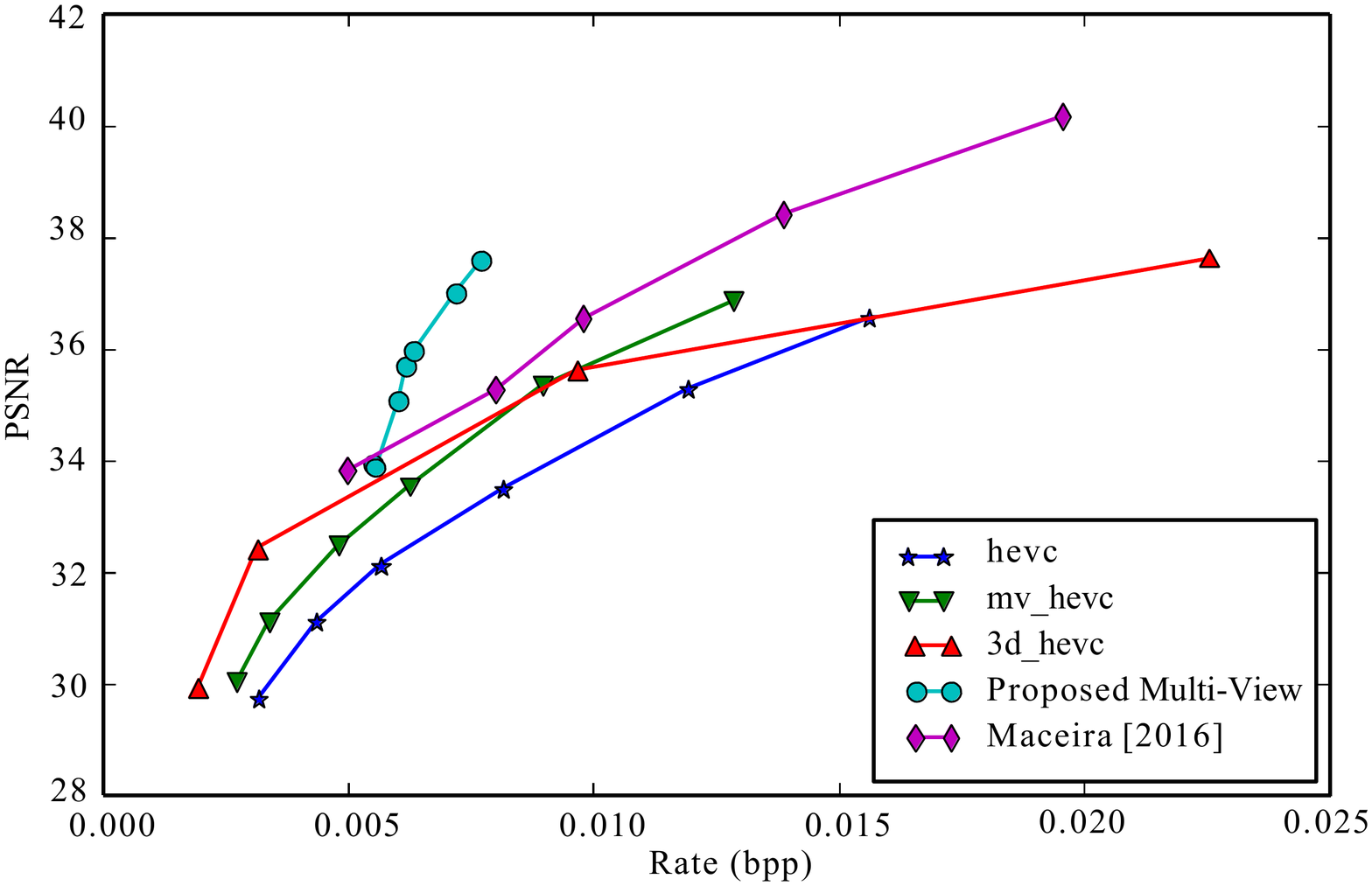}&
      		\includegraphics[trim=35 0 35 0,clip,width=0.46\textwidth]{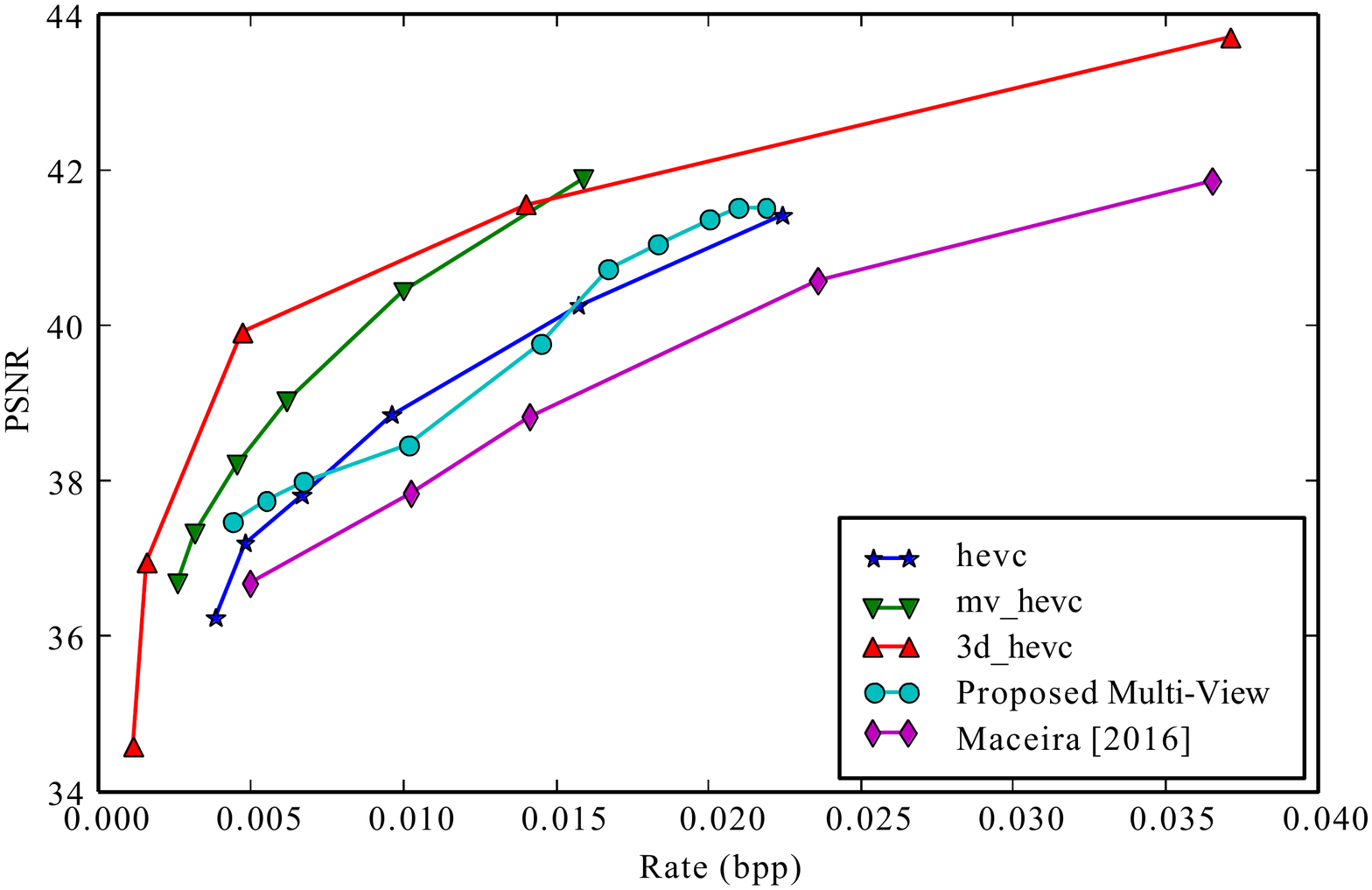}\\
	     	a) \emph{Undo Dancer} & b) \emph{Ghost Town}\\
      		\includegraphics[trim=35 0 35 0,clip,width=0.46\textwidth]{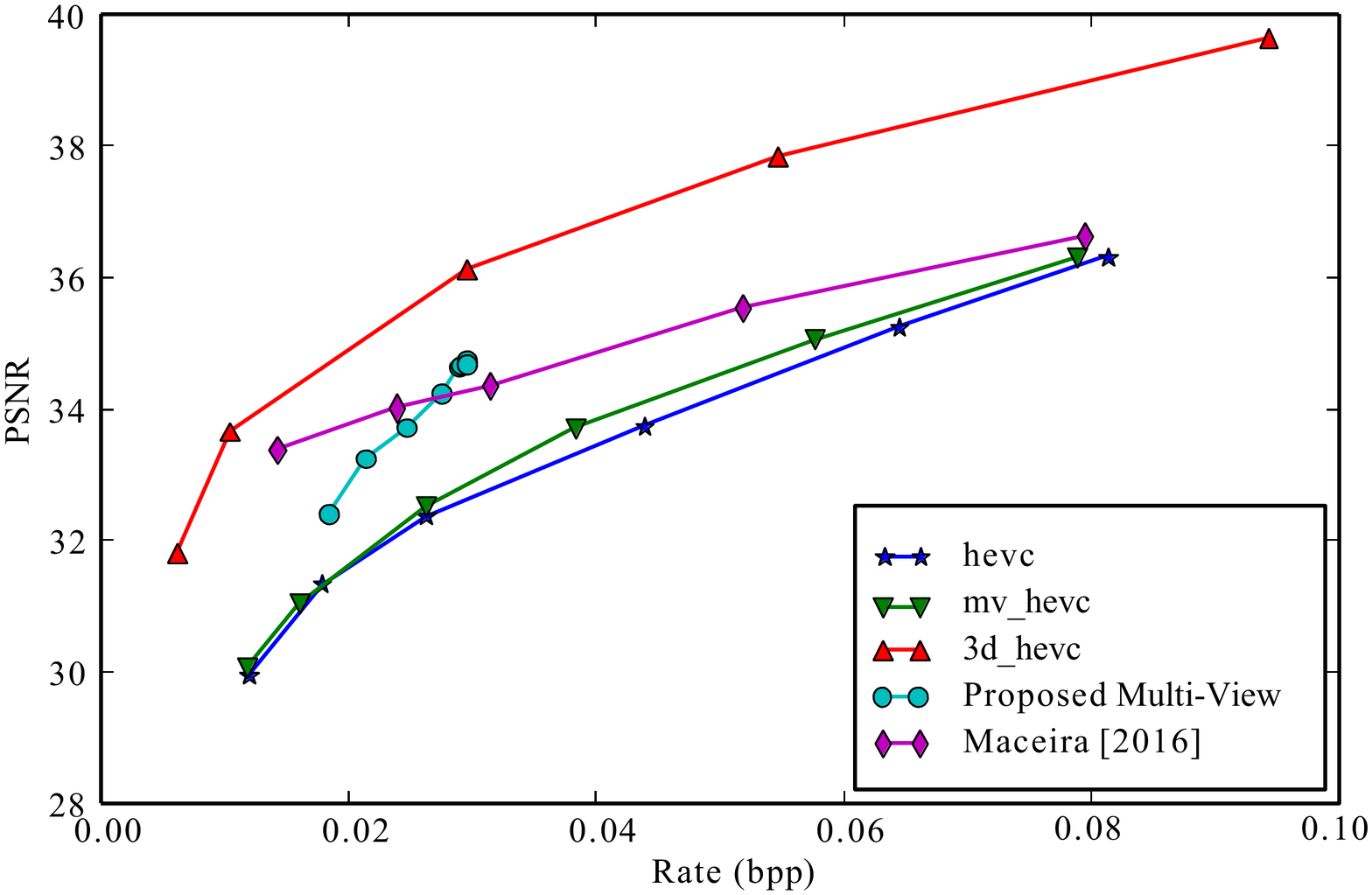}&   
      		\includegraphics[trim=35 0 35 0,clip,width=0.46\textwidth]{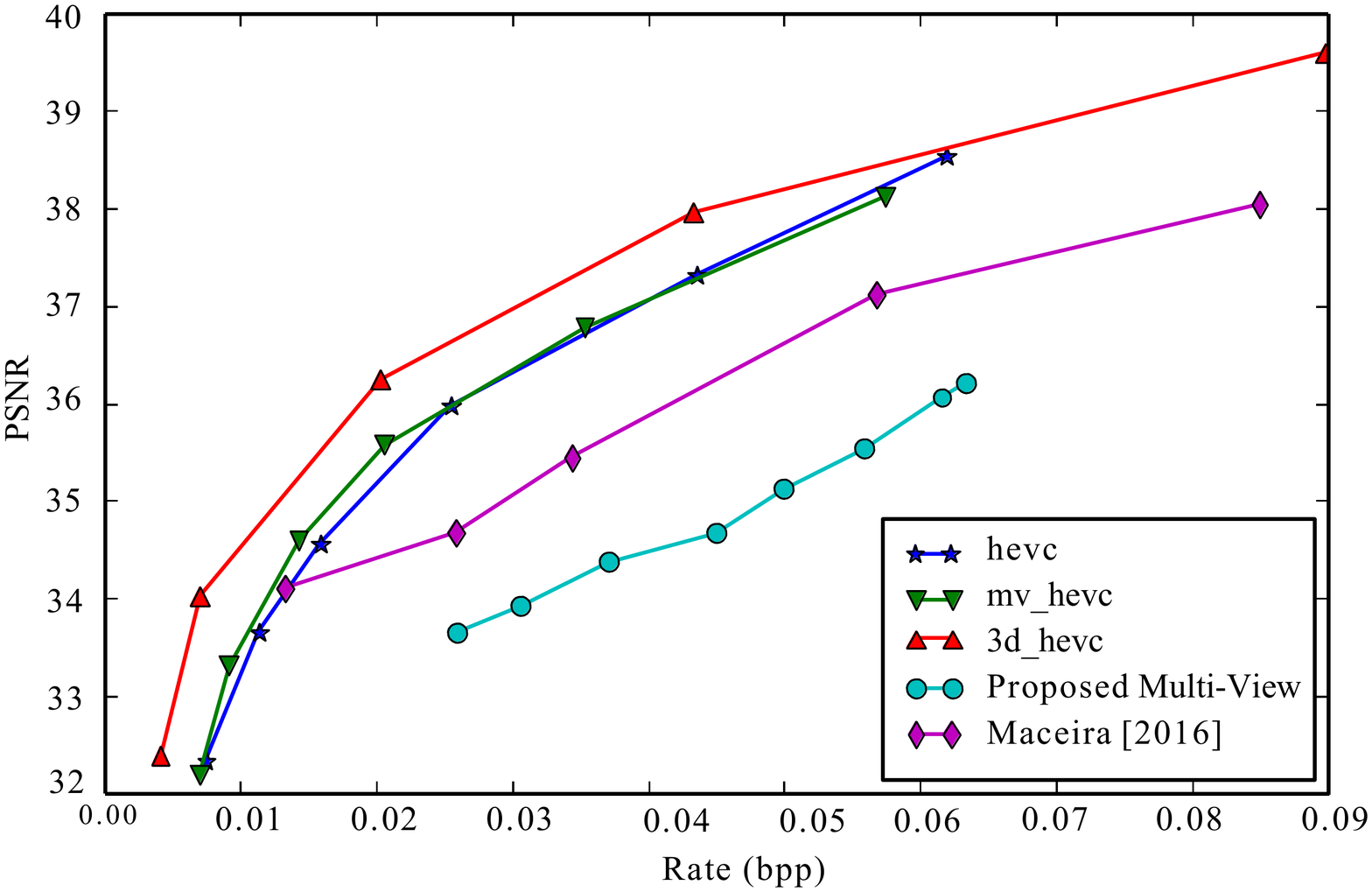}\\
	   	c) \emph{Ballet} & d) \emph{Breakdancers}\\
    \end{tabular}
  \end{center}
  \caption{Rate distortion results evaluated over rendered views.}
  \label{fig:rate_dist_render_orig}
\end{figure*}

Results measured on the depth map for the different sequences are shown in \figref{fig:rate_dist_depth_map}. In that comparison, the \emph{3D-HEVC} performs worse than the other \emph{HEVC} configurations methods of the literature. This is because color and depth map for two views are encoded together in \emph{3D-HEVC}, the view synthesis optimization maximize the quality in the virtual view and not directly in the depth map. The proposed method outperforms~\cite{Maceira2016} ([Maceira2016]) for all the sequences. The use of multi-view redundancy as well as the proposed optimization net allows a more than 2 dB gain for the different sequences. This improvement is noteworthy for the \emph{Undo Dancer} sequence, where the planar characteristic of the scene can be fully exploited. On the other hand, our method has more problems with the \emph{Ghost Town} sequence, where the number of added depth contours is similar to~\cite{Maceira2016}, thus achieving similar results. This is because the color segmentation is not able to properly segment elements at different depths since all of them have the same tonality.

In sequences where the depth map is noisier, as \emph{Ballet} and \emph{Breakdancers}, the 3D planar segments are not able to represent correctly the depth maps, thus obtaining worse results than the \emph{HEVC} standards.

Results in the rendered virtual view are shown in \figref{fig:rate_dist_render_orig}. The results of the planar implementations achieve better results than \emph{HEVC}  and \emph{MV-HEVC} in \emph{Ballet} and \emph{Undo Dancer} and are closer than in the depth map comparison in the other sequences. Notice that in \emph{Undo Dancer} \emph{3D-HEVC} does not improve \emph{HEVC} or \emph{MV-HEVC}. This is due to \emph{3D-HEVC} encoding together color and depth information. Here we only take the depth maps from the \emph{3D-HEVC}, using the same color image than the other methods for doing the rendered view. The results for \figref{fig:rate_dist_render_orig} are summarized in Tables 1 and 2 where the Bjontegaard's metric to compute the average gain in psnr (BD-SNR) and to compute the average saving in bitrate (BD-Rate) are shown, taking the \emph{HEVC} as a reference.

\begin{table}[ht]
\centering
\caption{BD-RATE}
\label{bd-rate}
\begin{tabular}{lcccc}
 & MV-HEVC & 3D-HEVC & Maceira2016 & Proposed \\
Undo Dancer & -23.69 & -40.97 & -36.94 & -49.16 \\ 
Ghost Town & -39.96 & -64.50 & 42.03 & 2.91 \\ 
Ballet & -6.03 & -71.89 & -37.16 & -41.63 \\ 
Breakdancers & -6.16 & -35.71 & 48.46 & 150.13 \\ 
\end{tabular} \\ 
\end{table}
\begin{table}[ht]
\centering
\caption{BD-SNR}
\label{bd.snr}
\begin{tabular}{lcccc}
 & MV-HEVC & 3D-HEVC & Maceira2016 & Proposed \\ 
Undo Dancer & 1.16 & 1.80 & 2.12 & 3.38 \\ 
Ghost Town & 1.43 & 2.16 & -1.03 & 0.01 \\ 
Ballet & 0.20 & 3.47 & 1.57 & 1.51 \\ 
Breakdancers & 0.17 & 1.02 & -1.03 & -2.57 \\ 
\end{tabular} \\ 
\end{table}

\replaced[id=mm]{The multi-view method proposed improves the Rate-Distortion results of \cite{Maceira2016} for all sequences, achieving encoding results competitive with the HEVC standards. However, this result is not translated equally in virtual views. For instance, \emph{Ballet} and \emph{Breakdancers} sequences are noisier and present a larger baseline between views. In this case, the proposed method reduces the number of 3D planes when encoding the corresponding depth maps, which penalizes the performance of the method. On the other hand, in sequence \emph{Undo Dancer}, the planar characteristics of their depth maps allow an improvement over the HEVC standards. In addition, the multi-view method presented in this work extracts a consistent segmentation for the different views, which could be very useful in further applications.}{The proposed method improves the Rate-Distortion results of Maceira2016 for all the sequences in the depth map. This result is not translated equally to the virtual view. For \emph{Ballet} and \emph{Breakdancers}, the results in the virtual view are below our proposal for encoding the depth map of one view. This is because our algorithm reduces the number of 3D planes when encoding the depth map. \emph{Ballet} and \emph{Breakdancers} are noisier and present a larger base-line between views. The reduced number of segments in the multi-view scheme penalizes the performance of the method. On the other hand, in \emph{Undo Dancer} the planar characteristics of their depth maps allow an improvement over the HEVC standards.}



\section{Conclusion}\label{sec:conclusion}


In this work we have presented a region-based multi-view depth map coding technique \replaced[id=rm]{based on a global 3D scene representation. Color and depth segmentations of the different views are combined into a single hierarchical representation. 
An optimization process over this hierarchy allows to obtain the optimal coding partition. This representation retrieves a unique 3D planar decomposition of the scene. A method to encode the 3D planar decomposition for multi-view depth map coding application is also proposed. Texture information is sent by modeling each resulting region as a 3D plane and sending the corresponding plane parameters. The use of a color partition (that is already available at the decoder, without extra cost) allows to reduce the cost of sending this coding partition. The multi-view coding method  shows competitive results against \emph{HEVC}.}{The proposed depth map coding technique combines a color partition and a depth map partition to obtain the final coding partition without encoding all depth edges. We propose a hierarchical optimization in terms
of rate distortion to find a consistent segmentation among the different views. This representation retrieves a unique 3D planar decomposition of the scene. A method to encode the 3D planar decomposition for depth map coding application is also proposed that shows competitive results against HEVC standards.}

\added[id=rm]{A side benefit of the proposed method is that, in addition to coding, the global 3D scene representation is susceptible to be used in tasks such as action detection~\cite{Zheng2015}, scene recognition~\cite{Gupta2013} or scene labeling~\cite{Wang2015}.}

\begin{acknowledgements}

This work has been developed in the framework of projects TEC2013-43935-R and TEC2016-75976-R, financed by the Spanish Ministerio de Econom\'ia y Competitividad and the European Regional Development Fund (ERDF)

\end{acknowledgements}

\bibliographystyle{spbasic}      
\bibliography{mv-coding-mmaceira}

\begin{thebibliography}{36}
\providecommand{\natexlab}[1]{#1}
\providecommand{\url}[1]{{#1}}
\providecommand{\urlprefix}{URL }
\expandafter\ifx\csname urlstyle\endcsname\relax
  \providecommand{\doi}[1]{DOI~\discretionary{}{}{}#1}\else
  \providecommand{\doi}{DOI~\discretionary{}{}{}\begingroup
  \urlstyle{rm}\Url}\fi
\providecommand{\eprint}[2][]{\url{#2}}

\bibitem[{Barrera and Padoy(2014)}]{Barrera2014}
Barrera F, Padoy N (2014) Piecewise planar decomposition of {3D} point clouds
  obtained from multiple static rgb-d cameras. In: 2014 2nd International
  Conference on {3D} Vision, vol~1, pp 194--201

\bibitem[{Charikar et~al(2003)Charikar, Guruswami, and Wirth}]{Charikar2003}
Charikar M, Guruswami V, Wirth A (2003) Clustering with qualitative
  information. In: Proceedings of the 44th Annual IEEE Symposium on Foundations
  of Computer Science, IEEE Computer Society, Washington, DC, USA, FOCS '03, pp
  524--533

\bibitem[{Fehn(2004)}]{Fehn2004}
Fehn C (2004) Depth-image-based rendering ({DIBR}), compression, and
  transmission for a new approach on {3D-TV}

\bibitem[{Fischler and Bolles(1981)}]{Fischler1981}
Fischler MA, Bolles RC (1981) Random sample consensus: A paradigm for model
  fitting with applications to image analysis and automated cartography. Commun
  ACM 24(6):381--395

\bibitem[{Freeman(1961)}]{Freeman1961}
Freeman H (1961) On the encoding of arbitrary geometric configurations. IRE
  Transactions on Electronic Computers EC-10(2):260--268

\bibitem[{Gao et~al(2016)Gao, Cheung, Maugey, Frossard, and Liang}]{Gao2016}
Gao Y, Cheung G, Maugey T, Frossard P, Liang J (2016) Encoder-driven inpainting
  strategy in multiview video compression. IEEE Transactions on Image
  Processing 25(1):134--149

\bibitem[{Glasner et~al(2011)Glasner, Vitaladevuni, and Basri}]{Glasner2011}
Glasner D, Vitaladevuni SN, Basri R (2011) Contour-based joint clustering of
  multiple segmentations. In: Proceedings of the 2011 IEEE Conference on
  Computer Vision and Pattern Recognition, IEEE Computer Society, Washington,
  DC, USA, CVPR '11, pp 2385--2392

\bibitem[{Gupta et~al(2013)Gupta, Arbel\'{a}ez, and Malik}]{Gupta2013}
Gupta S, Arbel\'{a}ez P, Malik J (2013) Perceptual organization and recognition
  of indoor scenes from {RGB-D} images. In: Computer Vision and Pattern
  Recognition (CVPR), 2013 IEEE Conference on, pp 564--571

\bibitem[{Kowdle et~al(2012)Kowdle, Sinha, and Szeliski}]{Kowdle2012}
Kowdle A, Sinha S, Szeliski R (2012) Multiple view object cosegmentation using
  appearance and stereo cues. In: European Conference on Computer Vision,
  Firenze, Italy, pp 789--803, \doi{10.1007/978-3-642-33715-4\_57}

\bibitem[{Liang and Zheng(2015)}]{Zheng2015}
Liang B, Zheng L (2015) A survey on human action recognition using depth
  sensors. In: Digital Image Computing: Techniques and Applications (DICTA),
  2015 International Conference on, pp 1--8

\bibitem[{Lucas et~al(2015)Lucas, Wegner, Rodrigues, Pagliari, da~Silva, and
  de~Faria}]{Lucas2015}
Lucas LFR, Wegner K, Rodrigues NMM, Pagliari CL, da~Silva EAB, de~Faria SMM
  (2015) Intra predictive depth map coding using flexible block partitioning.
  IEEE Transactions on Image Processing 24(11):4055--4068

\bibitem[{Maceira et~al(2016)Maceira, Morros, and Ruiz-Hidalgo}]{Maceira2016}
Maceira M, Morros JR, Ruiz-Hidalgo J (2016) Depth map compression via {3D}
  region-based representation. Multimedia Tools and Applications pp 1--24

\bibitem[{Merkle et~al(2007)Merkle, Smolic, Muller, and Wiegand}]{Merkle2007}
Merkle P, Smolic A, Muller K, Wiegand T (2007) Efficient prediction structures
  for multiview video coding. IEEE Transactions on Circuits and Systems for
  Video Technology 17(11):1461--1473

\bibitem[{Merkle et~al(2016)Merkle, M\"{u}ller, Marpe, and
  Wiegand}]{Merkle2016}
Merkle P, M\"{u}ller K, Marpe D, Wiegand T (2016) Depth intra coding for {3D}
  video based on geometric primitives. IEEE Transactions on Circuits and
  Systems for Video Technology 26(3):570--582

\bibitem[{Micusik and Kosecka(2009)}]{Micusik2009}
Micusik B, Kosecka J (2009) Piecewise planar city {3D} modeling from street
  view panoramic sequences. In: Computer Vision and Pattern Recognition, 2009.
  CVPR 2009. IEEE Conference on, pp 2906--2912

\bibitem[{M\"{u}ller et~al(2011)M\"{u}ller, Merkle, and Wiegand}]{Muller2011}
M\"{u}ller K, Merkle P, Wiegand T (2011) {3-D} video representation using depth
  maps. Proceedings of the IEEE 99(4):643--656

\bibitem[{M\"{u}ller et~al(2013)M\"{u}ller, Schwarz, Marpe, Bartnik, Bosse,
  Brust, Hinz, Lakshman, Merkle, Rhee, Tech, Winken, and Wiegand}]{Muller2013}
M\"{u}ller K, Schwarz H, Marpe D, Bartnik C, Bosse S, Brust H, Hinz T, Lakshman
  H, Merkle P, Rhee FH, Tech G, Winken M, Wiegand T (2013) {3D}
  {H}igh-{E}fficiency {V}ideo coding for multi-view video and depth data. IEEE
  Transactions on Image Processing 22(9):3366--3378

\bibitem[{Ortega and Ramchandran(1998)}]{Ortega98}
Ortega A, Ramchandran K (1998) Rate-distortion methods for image and video
  compression. IEEE Signal Processing Magazine 15(6):23--50

\bibitem[{Ostermann et~al(2004)Ostermann, Bormans, List, Marpe, Narroschke,
  Pereira, Stockhammer, and Wedi}]{Ostermann2004}
Ostermann J, Bormans J, List P, Marpe D, Narroschke M, Pereira F, Stockhammer
  T, Wedi T (2004) Video coding with {H.264/AVC}: tools, performance, and
  complexity. IEEE Circuits and Systems Magazine 4(1):7--28

\bibitem[{\"{O}zkalaycı and Alatan(2014)}]{Ozkalayci2014}
\"{O}zkalaycı BO, Alatan AA (2014) {3D} planar representation of stereo depth
  images for {3DTV} applications. IEEE Transactions on Image Processing
  23(12):5222--5232

\bibitem[{Ren et~al(2012)Ren, Bo, and Fox}]{Ren2012}
Ren X, Bo L, Fox D (2012) {RGB-(D)} scene labeling: Features and algorithms.
  In: Computer Vision and Pattern Recognition (CVPR), 2012 IEEE Conference on,
  pp 2759--2766

\bibitem[{Rusanovskyy et~al(2011)Rusanovskyy, Aflaki, and
  Hannuksela}]{rusanovskyy2011undo}
Rusanovskyy D, Aflaki P, Hannuksela M (2011) Undo dancer {3DV} sequence for
  purposes of {3DV} standardization. ISO/IEC JTC1/SC29/WG11 MPEG2010 M 20028

\bibitem[{Salembier and Garrido(2000)}]{Salembier2000}
Salembier P, Garrido L (2000) Binary partition tree as an efficient
  representation for image processing, segmentation, and information retrieval.
  IEEE Transactions on Image Processing 9(4):561--576

\bibitem[{Schwarz et~al(2011)Schwarz, Mateus, Lallemand, and
  Navab}]{Schwarz2011}
Schwarz LA, Mateus D, Lallemand J, Navab N (2011) Tracking planes with time of
  flight cameras and j-linkage. In: Applications of Computer Vision (WACV),
  2011 IEEE Workshop on, pp 664--671

\bibitem[{Shoham and Gersho(1988)}]{Shoham1988}
Shoham Y, Gersho A (1988) Efficient bit allocation for an arbitrary set of
  quantizers 36(9):1445--1453

\bibitem[{Silberman et~al(2012)Silberman, Hoiem, Kohli, and
  Fergus}]{Silberman2012}
Silberman N, Hoiem D, Kohli P, Fergus R (2012) Indoor segmentation and support
  inference from {RGBD} images. In: ECCV

\bibitem[{Sinha et~al(2009)Sinha, Steedly, and Szeliski}]{Sinha2009}
Sinha S, Steedly D, Szeliski R (2009) Piecewise planar stereo for image-based
  rendering. In: International Conference on Computer Vision, Kyoto, Japan, pp
  1881--1888

\bibitem[{Sullivan et~al(2012)Sullivan, Ohm, Han, and Wiegand}]{Sullivan2012}
Sullivan GJ, Ohm JR, Han WJ, Wiegand T (2012) Overview of the high efficiency
  video coding ({HEVC}) standard. IEEE Transactions on Circuits and Systems for
  Video Technology 22(12):1649--1668

\bibitem[{Sullivan et~al(2013)Sullivan, Boyce, Chen, Ohm, Segall, and
  Vetro}]{Sullivan2013}
Sullivan GJ, Boyce JM, Chen Y, Ohm JR, Segall CA, Vetro A (2013) Standardized
  extensions of {H}igh {E}fficiency {V}ideo coding ({HEVC}). IEEE Journal of
  Selected Topics in Signal Processing 7(6):1001--1016

\bibitem[{Torres and Kunt(1996)}]{Torres1996}
Torres L, Kunt M (1996) Second Generation Video Coding Techniques, Springer US,
  Boston, MA, pp 1--30

\bibitem[{Varas et~al(2015)Varas, Alfaro, and Marques}]{Varas2015}
Varas D, Alfaro M, Marques F (2015) Multiresolution hierarchy co-clustering for
  semantic segmentation in sequences with small variations. In: 2015 IEEE
  International Conference on Computer Vision (ICCV), pp 4579--4587

\bibitem[{Verleysen and De~Vleeschouwer(2016)}]{Verleysen2016}
Verleysen C, De~Vleeschouwer C (2016) Piecewise-planar 3d approximation from
  wide-baseline stereo. In: The IEEE Conference on Computer Vision and Pattern
  Recognition (CVPR)

\bibitem[{Wang et~al(2015)Wang, Lu, Cai, Wang, and Cham}]{Wang2015}
Wang A, Lu J, Cai J, Wang G, Cham TJ (2015) Unsupervised joint feature learning
  and encoding for {RGB-D} scene labeling. IEEE Transactions on Image
  Processing 24(11):4459--4473

\bibitem[{Yin et~al(2015)Yin, Velastin, Ellis, and Makris}]{Yin2015}
Yin F, Velastin SA, Ellis T, Makris D (2015) Learning multi-planar scene models
  in multi-camera videos. IET Computer Vision 9(1):25--40

\bibitem[{Zhang et~al(2011)Zhang, Li, Li, Rusanovskyy, and
  Hannuksela}]{zhang2011ghost}
Zhang J, Li R, Li H, Rusanovskyy D, Hannuksela MM (2011) Ghost {T}own {F}ly
  {3DV} sequence for purposes of {3DV} standardization. ISO/IEC JTC1/SC29/WG11,
  Doc M 20027

\bibitem[{Zitnick et~al(2004)Zitnick, Kang, Uyttendaele, Winder, and
  Szeliski}]{Zitnick2004}
Zitnick CL, Kang SB, Uyttendaele M, Winder S, Szeliski R (2004) High-quality
  video view interpolation using a layered representation. In: ACM SIGGRAPH
  2004 Papers, New York, NY, USA, SIGGRAPH '04, pp 600--608

\end{thebibliography}

%
%

\end{document}